\documentclass[10pt,twocolumn,letterpaper]{article}

\usepackage[pagenumbers]{cvpr} 

\usepackage[dvipsnames]{xcolor}

\definecolor{cvprblue}{rgb}{0.21,0.49,0.74}
\usepackage[pagebackref,breaklinks,colorlinks,citecolor=cvprblue]{hyperref}
\usepackage{comment}
\usepackage{xcolor}
\usepackage{graphicx}
\usepackage{amsmath}
\usepackage{amssymb}
\usepackage{booktabs}

\usepackage{subcaption}
\usepackage{xcolor}
\usepackage{cuted}
\usepackage{etoolbox}
\usepackage{bbm}
\usepackage{mathrsfs}
\usepackage{color, colortbl}

\newif\ifdraft
\drafttrue

\def\confName{CVPR}
\def\confYear{2024}

\title{Learning Feature-Preserving Portrait Editing from Generated Pairs}

\author{Bowei Chen$^{1}$\thanks{Work done during internship at ByteDance.} \qquad Tiancheng Zhi$^{2}$ \qquad  Peihao Zhu$^{2}$ \qquad Shen Sang$^{2}$ \qquad Jing Liu$^{2}$ \qquad Linjie Luo$^{2}$ \\
$^1$ University of Washington, $^2$ ByteDance \\
boweiche@cs.washington.edu, \\
\{tiancheng.zhi, peihao.zhu, shen.sang, jing.liu, linjie.luo\}@bytedance.com}

\begin{document}

\maketitle

\begin{strip}
\begin{minipage}{\textwidth}\centering
\vspace{-10pt}
    \includegraphics[width=\textwidth]{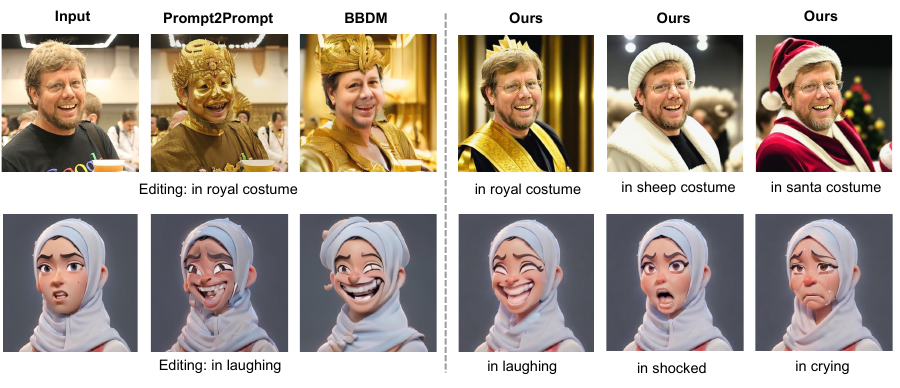}
\captionof{figure}{Our method takes a portrait image as input, and applies advanced editing effects with our proposed framework. We can handle both real human portraits (1st row) as well as cartoon characters (2nd row). Our approach obtains superior aesthetic quality while at the same time preserving key features from the input subject. Compared with baseline approaches (left), we achieve better subject feature preservation (\eg, identity), structural alignment, and fewer artifacts.}
\label{fig:teaser}
\end{minipage}
\end{strip}


\begin{abstract}
Portrait editing is challenging for existing techniques due to difficulties in preserving subject features like identity.
In this paper, we propose a training-based method leveraging auto-generated paired data to learn desired editing while ensuring the preservation of unchanged subject features. 
Specifically, we design a data generation process to create reasonably good training pairs for desired editing at low cost. 
Based on these pairs, we introduce a Multi-Conditioned Diffusion Model to effectively learn the editing direction and preserve subject features. 
During inference, our model produces accurate editing mask that can guide the inference process to further preserve detailed subject features.  
Experiments on costume editing and cartoon expression editing show that our method achieves state-of-the-art quality, quantitatively and qualitatively.
\end{abstract}

\section{Introduction}
\label{sec:intro}

Portrait editing is increasingly favored in photo and social applications. 
In many of these applications, users can select from a set of pre-defined editing options and then apply their chosen edits to their own photos.
In practice, the key requirement of portrait editing is to deliver outcomes that achieve selected editing while \textit{strictly} preserving the features of subjects intended to remain unaltered (\eg, identity and clothing for expression editing). 
Nevertheless, meeting this requirement poses a considerable challenge, as even slight deviations in these features can markedly affect the perceived quality of the outcome. 
Therefore, the goal of this paper is to design a portrait editing pipeline that can achieve superior editing outcomes for a specific editing task favored by users.

Existing image editing approaches fail to satisfy the requirements of portrait editing tasks.
They can be categorized into two types.
The first one is training-free methods, which mainly rely on a pretrained diffusion model~\cite{rombach2021highresolution} to perform editing guided by a text prompt. However, they suffer from two limitations.
(1) They struggle to achieve desired editing as they depend on inversion techniques to reverse the input image into a denoising process, which may hurt editability.
(2) They fail to preserve detailed subject features as little prior knowledge for invariance is enforced.  
Figure \ref{fig:teaser} (left) shows outputs of a training-free method Prompt2Prompt~\cite{hertz2022prompt}.
Another stream of work is training-based methods, aiming to learn the editing direction for desired changes, and also preserve untargeted subject features, with a training set.
However, these methods require extremely high-quality training dataset, which is usually hard to collect.  
Figure \ref{fig:teaser} (left) shows outputs of a recent training-based method BBDM~\cite{li2023bbdm}.

In this paper, we opt for training using a synthetic dataset generated automatically at low cost, thereby eliminating the necessity of manually collecting datasets. 
Our framework generates a synthetic dataset for any user-defined editings and uses this dataset to effectively learn the editing directions, fulfilling the aforementioned requirements, and upholding high image quality.
Specifically,  we first design a conditional dataset generation strategy to produce diverse paired data given text prompts, which has better identity and layout alignment than existing data generation strategy. 
Given these pairs, we design a Multi-Conditioned Diffusion Model (MCDM) to effectively learn editing direction and preserve the subject features.  
This is achieved by injecting the conditional signals from input image and text prompt into the diffusion model through different ways. 
Finally, we demonstrate that the trained MCDM can \textit{explicitly} identify regions expected to change (\eg, face regions for expression editing), producing an editing mask.  This provides guidance for our inference process to further keep subject features untouched. 


As shown in Figure~\ref{fig:teaser}, our editing results achieve expressive styles while preserving subject features, in both real person costume and cartoon expression editing cases. The effectiveness of the method is further validated through comprehensive quantitative analysis and user studies, which collectively demonstrate its clear superiority over existing baseline methods.

\noindent\textbf{Contributions:} (1) A data generation technique providing paired data with better identity and layout alignment; (2) A Multi-Conditioned Diffusion Model producing feature-preserving results and accurate editing masks for inference guidance; (3) State-of-the-art portrait editing results.

\section{Related Work}
\label{sec:formatting}

Image generation and editing have seen significant advancements with generative models like GANs~\cite{NIPS2014_5ca3e9b1}, VAEs~\cite{kingma2013autoencoding}, and normalizing flows~\cite{rezende2015variational}, leading to highly realistic outputs~\cite{Karras_2019, Karras2019stylegan2}. Recent breakthroughs in diffusion models~\cite{ho2020denoising, sohl2015deep, song2019generative, song2020denoising}, such as Imagen~\cite{saharia2022photorealistic}, GLIDE~\cite{nichol2022glide}, DALL-E2~\cite{ramesh2022hierarchical}, and Stable Diffusion~\cite{rombach2021highresolution}, have further revolutionized this field. They can generate a wide variety of images from mere textual descriptions and has spurred research into their applications in image editing.

\noindent\textbf{Training-Free Approaches:} 
Prevalent editing methods rely on inverting images into a model's latent space~\cite{abdal2019image2stylegan,abdal2020image2stylegan++,wu2021stylespace,zhu2021improved,zhu2020indomain} and editing by manipulating latent codes~\cite{abdal2020image2stylegan++,shen2020interpreting, harkonen2020ganspace} or model weights~\cite{gal2021stylegan,roich2021pivotal,bau2020semantic,alaluf2022hyperstyle}, without new model training. They are known as training-free methods. Text-to-image diffusion models, akin to GANs, use Gaussian noise as latent input, combined with textual guidance, to generate images. Methods like SDEdit~\cite{meng2021sdedit} add noise to the input image for a fixed number of steps, and then initiate a text-guided denoising process for repainting. However, these methods apply global editing, failing to preserve details in areas not targeted for modification.
To overcome this issue, some studies~\cite{nichol2021glide, avrahami2022blended, avrahami2022blendedlatent} use user-provided masks to define editing regions, thus allowing for partial edits. Yet, obtaining precise masks for editing is non-trivial, and mask-based inpainting methods often result in the loss of image information within the masked area, disrupting the consistency between the pre- and post-edit images.

For controlled, local editing, Prompt2Prompt~\cite{hertz2022prompt} and DiffEdit~\cite{couairon2022diffedit} have been developed. The former preserves layout and subject geometry through cross-attention maps, while the latter generates an editing mask through contrasting predictions from different text conditions. Both methods employ DDIM inversion~\cite{song2020denoising,dhariwal2021diffusion} to encode input images. However, DDIM inversion, especially with classifier-free guidance, often leads to unsatisfactory reconstruction and editing outcomes. Null-text Inversion~\cite{mokady2023null} improves inversion reconstruction while retaining the editing capabilities. 
Pix2pix-zero~\cite{parmar2023zero} improves DDIM inversion through noise regularization~\cite{Karras2019stylegan2} and introduces cross-attention loss during the denoising process.
However, this method may pose difficulties in terms of control and could lead to unexpected outcomes, especially for portrait editing.

\noindent\textbf{Training-Based Approaches:} Training-based methods learn editing direction from a large dataset.
Li et al.~\cite{li2023bbdm}  and Sheynin et al.~\cite{sheynin2022knn} train diffusion models for image-to-image translation and local semantic editing without inversion, but their expressiveness and quality lag behind current large-scale diffusion models. InstructPix2Pix~\cite{Brooks_2023_CVPR} uses GPT-3~\cite{brown2020language} and Prompt2Prompt~\cite{hertz2022prompt} to create text edited pairs and distills a diffusion model, generally producing more controlled edits and showing robustness with real image inputs. The effectiveness of training-based methods depends on the quality of the constructed pairs. Our method, which falls into this category, achieves greater consistency and superior editing results by using Composable Diffusion~\cite{liu2022compositional} to generate better pairs. Relying on our condition injection mechanism and network design, we are capable of producing edits which are less affected by data imperfection, and thus better preserving input features.

Diffusion-based editing also relates to concept embedding~\cite{gal2022textual}, model fine-tuning~\cite{ruiz2022dreambooth}, and controlled generation~\cite{zhang2023adding,mou2023t2i}, but they are outside our discussion scope.

\section{Our Pipeline}
\label{sec:pipeline}

\begin{figure*}
    \centering
    \includegraphics[scale=0.68]{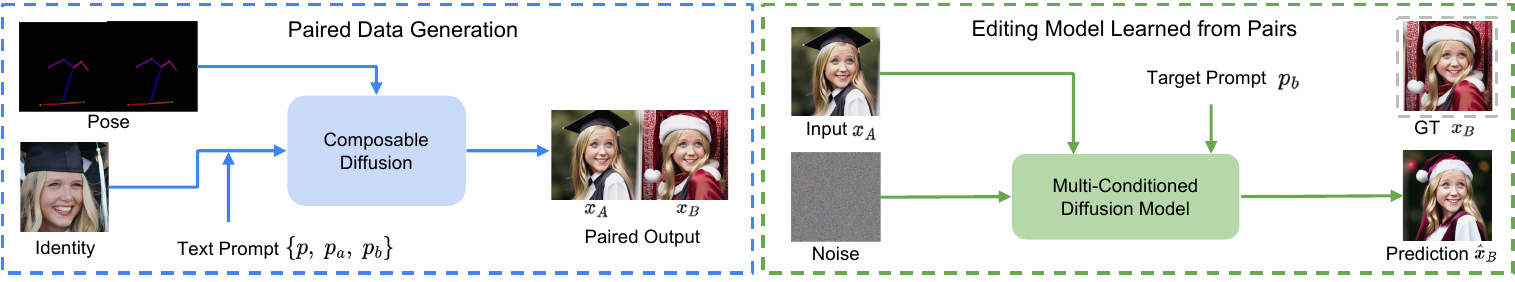}
    \caption{Overview of our pipeline.  \textit{Paired Data Generation} (blue dashed box) first constructs training pairs using Composable Diffusion~\cite{liu2022compositional} conditioning on pose and identity information.
    \textit{Multi-Conditioned Diffusion Model} (green dashed box) encodes multiple condition signals to learn the editing direction and preserve subject features based on the generated pairs.  The multi-condition design enhances the robustness in handling imperfections within training pairs.
    }
    \label{fig:overview}
\end{figure*}

Given an input portrait image $x_A$ in the source domain $A$, our goal is to synthesize a high-quality portrait image $\hat{x}_B$ in domain $B$. 
A well-edited image $\hat{x}_B$ should: 
(1) retain the untargeted subject features (\eg, identity) and rough layout from $x_A$,
(2) ensure editing fidelity (\ie, $\hat{x}_B \in B$) and maintain high image quality.

To this end, we design a diffusion-based image editing pipeline with three stages.
(1) 
We first introduce an automated data generation strategy to create reasonably good but not perfect pairs of input $x_A$ and ground truth $x_B$ (Figure \ref{fig:overview} left). 
(2) 
Then we design and train a Multi-Conditioned Diffusion Model (MCDM) (Figure \ref{fig:overview} right) on this generated dataset. 
By leveraging multiple conditions in different ways, MCDM can effectively learn the editing direction from the training pairs, while preserving detailed subject features that are not supposed to be changed. 
(3) 
During inference, we generate edited results using the trained MCDM with an automatically generated editing mask to further preserve subject details in $x_A$.


\subsection{Preliminary}

We start with a quick overview of Latent Diffusion~\cite{rombach2021highresolution} and establish notations that we use throughout.  
Latent Diffusion has two components: 
(1) a Variational Autoencoder, including an encoder $E$ to transform an image $x$ into a latent code $z = E(x)$, and a decoder $D$ to map $z$ back to an image $x'=D(z)$, 
(2) a U-Net $\epsilon_\theta (z_t, t, C)$ which predicts added noise given a noisy latent. $z_t$ is the noisy latent code at timestep $t$ and $C$ is a \textit{tuple} of conditional signals. 

To generate an image, a noisy latent  $z_T$ is randomly sampled and processed through denoising by the U-Net over a fixed number of timesteps, denoted as $T$. The iterative denoising process transforms $z_T$ into a clean latent $z_0$, which is subsequently utilized by the decoder $D$ to generate the image.
Specifically, at timestep $t$, the denoised latent $z_{t-1}$ is sampled based on $z_t$ and $\tilde{ \epsilon}_\theta (z_t, t, C)$, which is computed using classifier-free guidance~\cite{ho2022classifier}. Here is an example with two elements in $C$, given by:
\begin{align}
   \tilde{ \epsilon}_\theta (z_t, t, C) \nonumber 
    =&  \epsilon_\theta (z_t, t, \{\varnothing,\varnothing\})  \nonumber \\
+&  s_1 (\epsilon_\theta (z_t, t, \{c_1,\varnothing\}) - \epsilon_\theta (z_t, t, \{\varnothing,\varnothing\})) \nonumber \\
+&  s_2 (\epsilon_\theta (z_t, t, \{c_1,c_2\}) - \epsilon_\theta (z_t, t, \{c_1,\varnothing\})),
\label{eq:cfg}
\end{align}
where $c_1$ and $c_2$ denote two conditional signals, with $\varnothing$ representing a null value (\eg, a black image for image condition).  $s_1$ and $s_2$ are the weights for $c_1$ and $c_2$, respectively.  
Eq. \ref{eq:cfg} can also be easily rewritten to suit the case of one or three conditional signals.


For clarity, we primarily use the task of costume editing to illustrate the pipeline. The goal is to transform a person with a regular outfit into a Santa Claus costume.



\subsection{Paired Data Generation}
\label{subsec:dataset_generation}
The goal is to design a data generation strategy that can produce paired exemplars aligned with a specified editing direction (\eg, from regular to Santa Claus costumes) defined by text prompts.
However, generating pairs with perfect spatial and identity alignment is very challenging. 
Thus we seek to design a strategy (Figure \ref{fig:overview} left) that can generate reasonably good pairs, meeting these essential criteria: 
(1) the user identity in input $x_A$ and ground truth $x_B$ should match as closely as possible;
(2) $x_A$ and  $x_B$ should have rough spatial alignment;
(3) the data should cover a diverse range of user appearances (for better generalization).


\begin{figure}[!t]
\begin{center}
\begin{subfigure}[b]{0.49\linewidth}
\includegraphics[width=\linewidth]{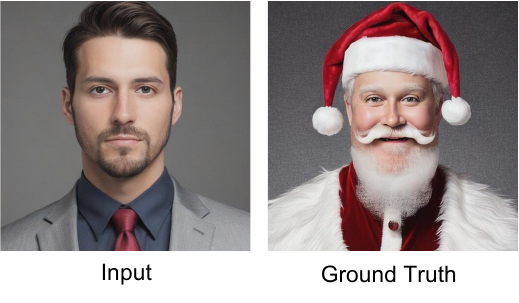} %
\caption{Prompt-to-Prompt Strategy}
\end{subfigure}
\begin{subfigure}[b]{0.49\linewidth}
\includegraphics[width=\linewidth]{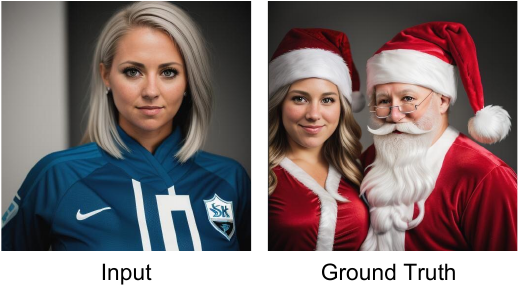} %
\caption{Our Strategy w/o Pose}
\end{subfigure}
\\
\vspace{2mm}
\begin{subfigure}[b]{0.49\linewidth}
\includegraphics[width=\linewidth]{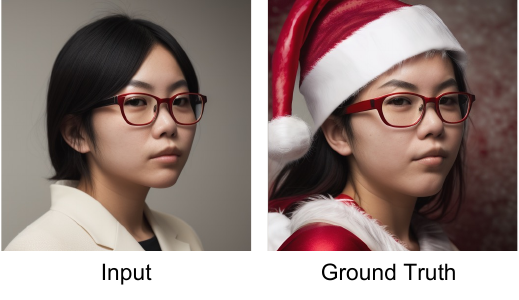} %
\caption{Our Strategy w/o ID}
\end{subfigure}
\begin{subfigure}[b]{0.49\linewidth}
\includegraphics[width=\linewidth]{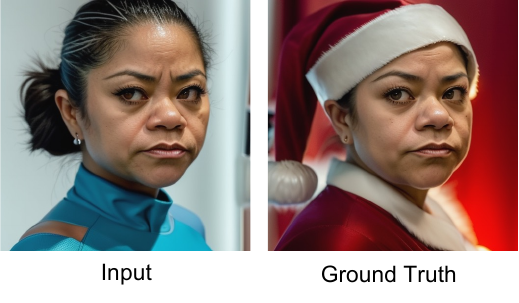} %
\caption{Our Strategy }
\end{subfigure}
\caption{Examples of pairs generated by different strategies. Prompt-to-Prompt (a) fails to produce pairs with consistent identity. 
Without pose condition, (b) produces pairs with significant spatial misalignment.
Without identity conditions, (c) results in pairs with obvious face shapes difference. Our strategy (d) significantly improves these issues. 
}
\label{fig:data_ablation}
 \ifdraft
\vspace{-5mm}
 \else
\fi
\end{center}
\end{figure}

One straightforward idea suggested by InstructPix2Pix~\cite{brooks2023instructpix2pix} is to use GPT-3~\cite{brown2020language} for generating a pair of text prompts in the source and target domains. These generated prompts are then employed to create $x_A$ and $x_B$ using a pretrained Stable Diffusion model~\cite{rombach2021highresolution} and the Prompt2Prompt image editing technique~\cite{hertz2022prompt}. However, this method often results in unsatisfactory $x_B$ as it fails to preserve the identity in $x_A$, as depicted in Figure~\ref{fig:data_ablation} (a).

Instead, we build a conditional pair generation strategy on top of Composable Diffusion~\cite{liu2022compositional} to meet the three requirements.  Key designs include: 
(1) Following ~\cite{liu2022compositional},  we generate $x_A$ and $x_B$ within a single image achieved through a single denoising process.  This helps generate consistent identities in $x_A$ and $x_B$ (criterion 1). 
(2) We incorporate pose information to improve spatial alignment (criterion 2). 
(3) We extract identity information from real photos and use this information to ensure criterion 1 and 3.

To implement design (1), we employ a pretrained Stable Diffusion in conjunction with Composable Diffusion~\cite{liu2022compositional} to generate an image $x = [x_A, x_B] \in \mathbb{R}^{H \times 2W \times 3}$, where the operator $[\cdot, \cdot]$ represents the horizontal concatenation of two images. Here, $H$ and $W$ denote the height and width of $x_A$ and $x_B$. 
Further,  the design (2) and (3) are implemented as conditions to guide the denoising process of $x$.

Specifically, we begin by randomly initializing a latent code $z_{T} \in \mathbb{R}^{h \times 2w \times 4}$, where $h=H/8$, $w=W/8$, and 4 represents the feature dimension of the latent code. 
At each timestep $t$, we compute the predicted noise by combining three classifier-free guidance results:
\begin{align}
    \bar\epsilon
    =& s'_d  \cdot  \tilde{\epsilon}_{\theta'} (z_{t}, t, \{c_{p}, c_{id}\}) \; \nonumber + \\& s'_a  \cdot  M'_a \odot \tilde{\epsilon}_{\theta'} (z_{t}, t, \{c_{p_a}, c_{id}\}) \;  \nonumber+ \\& s'_b  \cdot  M'_b  \odot  \tilde{\epsilon}_{\theta'} (z_{t}, t, \{c_{p_b}, c_{id}\}), 
    \label{eq:composable}
\end{align}
where $c_p$, $c_{p_a}$, and $c_{p_b}$ represent text embeddings computed from the shared prompt $p$, the source prompt $p_a$, and the target prompt $p_b$, respectively. 
In the example of Figure \ref{fig:overview}, $p$ is ``the same woman on the left and right'', $p_a$ is ``a woman, normal costume'', and $p_b$ is ``a woman, santa claus costume''. 
$c_{id}$ denotes identity embeddings (design (3)) extracted from a real-world portrait image using a variant of CLIP-based identity encoder~\cite{wei2023elite}, trained on the FFHQ dataset~\cite{Karras_2019}. This encoder translates an image into  multiple textual word embeddings, thus can be combined with $c_p$, $c_{p_a}$, and $c_{p_b}$ to provide identity information for the denoising process.
See supplementary for further details.

The matrices $M'_a$ and $M'_b$ are defined as $[\mathbf{1}, \mathbf{0}]$ and $[\mathbf{0}, \mathbf{1}]$ respectively, both belonging to $\mathbb{R}^{h \times 2w \times 4}$. Here, $\mathbf{1}$ ($\mathbf{0}$) represents a matrix in the dimension $h \times w \times 4$ with all values set to one (zero). Additionally, the variables $s'_d$, $s'_a$, and $s'_b$ signify the strengths associated with each predicted noise.
Furthermore, the denoising process is guided by a pose image (design (2)) using the OpenPose~\cite{cao2017realtime} ControlNet~\cite{zhang2023adding}, as shown in Figure \ref{fig:overview} top left. This pose image ensures alignment by featuring the same pose in both the left and right parts of the image. The pair generated by our approach is depicted in Figure \ref{fig:overview} on the left.

\begin{figure}[!t]
\begin{center}
\begin{subfigure}[b]{0.32\linewidth}
\includegraphics[width=\linewidth]{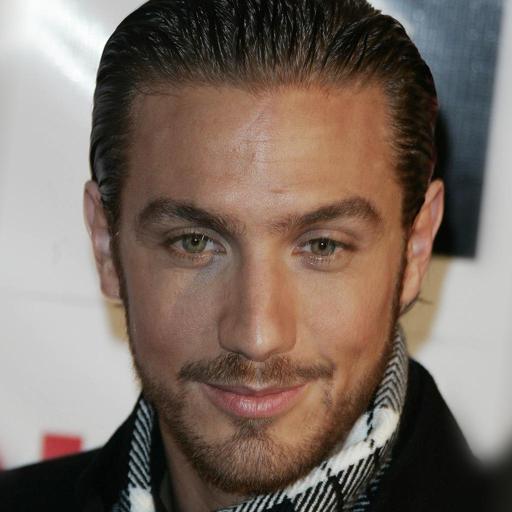} %
\caption{Input}
\end{subfigure}
\begin{subfigure}[b]{0.32\linewidth}
\includegraphics[width=\linewidth]{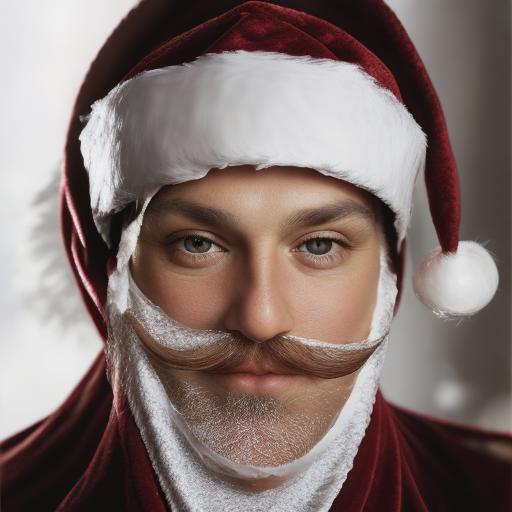} %
\caption{ Output w/o ID}
\end{subfigure}
\begin{subfigure}[b]{0.32\linewidth}
\includegraphics[width=\linewidth]{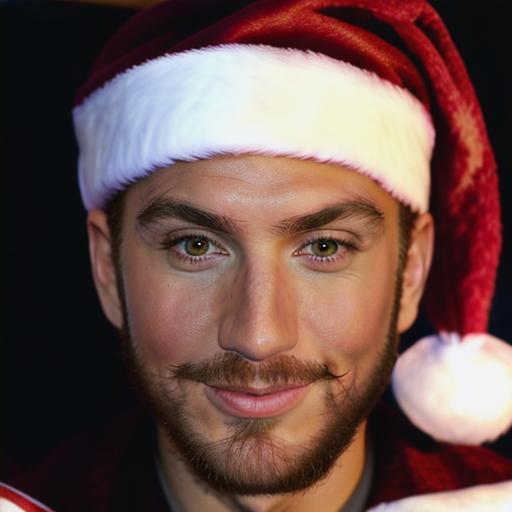} %
\caption{ Output with ID}
\end{subfigure}
\caption{  Training on a dataset with less diverse identities (b) results in inconsistent identity with the input (a). Conversely, training on a dataset with diverse identities yields the desired editing outcome (c), demonstrating its better generalization ability.
}
\label{fig:id_ablation}
\end{center}
\end{figure}

Notably, both design (2) (for pose) and design (3) (for identity) play a crucial role in generating good pairs.  Figure \ref{fig:data_ablation} illustrates this point.
Dropping one of them results in considerable spatial misalignment (b) and noticeable differences in facial shape (c). 
In addition, design (3) also contributes to generating diverse individuals across different pairs. This is crucial for enhancing generalization ability, as shown in Figure \ref{fig:id_ablation}.


\subsection{Training Multi-Conditioned  Diffusion Model}

Although the generated pairs are reasonably good, they are still not perfect. 
For example, in Figure \ref{fig:overview}, the face in $x_B$ is slightly wider than that in $x_A$. 
The imperfection can potentially confuse the model and harm the performance.

\begin{figure}
    \centering
    \includegraphics[scale=0.7]{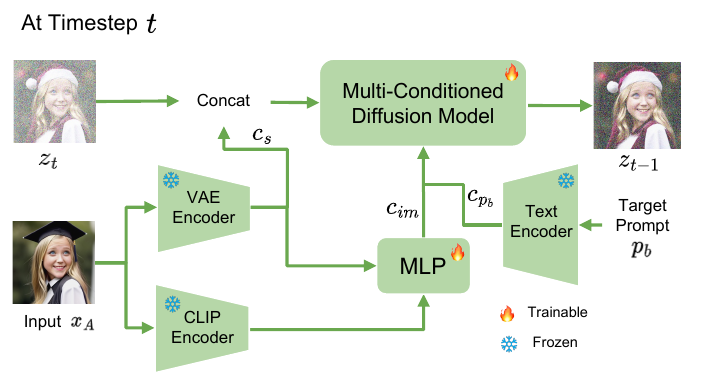}
    \caption{Illustration of Multi-Conditioned Diffusion Model, where both image and text embeddings are injected into the model through different ways to effectively learn the editing direction and preserve subject features.
    }
     \ifdraft
\vspace{-2mm}
 \else
\fi
    \label{fig:mcdm}
\end{figure}

Therefore, given these imperfect pairs, we design an image editing model to effectively learn pertinent information, such as editing direction and preservation of untargeted subject features, from the generated pairs while simultaneously filtering out unexpected noise -- specifically, small variations in identity and layout. 
Inspired by ~\cite{karras2023dreampose}, the key design of our model is to integrate various conditions into the Stable Diffusion architecture in distinct ways. We call our model Multi-Conditioned Diffusion Model (MCDM).
We will first define these conditions, and later elaborate how they help learn  pertinent information from imperfect data through different injection ways. 
The details of the MCDM are shown in Figure \ref{fig:mcdm}.

Our model $\epsilon_{\theta} (z_t, t,  \{c_s, c_{im}, c_{p_b}\})$ considers three pathways of conditional signals: 
 (1) spatial embeddings $c_s = E(x_A)$,
(2) text embeddings $c_{p_b}$, extracted by pretrained Stable Diffusion text encoder with target text prompt $p_b$ as input,
 (3) image embeddings $c_{im} = MLP([E(x_A), CLIP_{im}(x_A)]$), where $CLIP_{im}(\cdot)$ denotes embeddings extracted from the pretrained CLIP image encoder~\cite{radford2021learning}. $MLP(\cdot)$ is a multi-layer perceptron that projects image embeddings to the space of text embeddings. 

To incorporate these embeddings into our model, we make modifications to the Stable Diffusion architecture as follows.
(1) To prevent the imperfections in $x_B$ from misleading the model into generating an output $\hat{x}_B$ that alters the layout and identity in $x_A$, we concatenate the spatial embeddings $c_s$ with the noisy latent $z_t$ (input of U-Net).
The resulting concatenation is then utilized as the input for the U-Net. Architecturally, the first layer of the U-Net encoder is adjusted to accommodate an additional 4 channels (for $c_s$), increasing the total to 8 channels.
(2) $c_{p_b}$ and $c_{im}$ are concatenated and fed into the cross-attention layer, akin to the Stable Diffusion architecture.  
Functionally, $c_{p_b}$ includes crucial information about the target domain as instructed by the text prompt, steering the output $\hat{x}_B$ towards the desired domain $B$. Simultaneously, $c_{im}$ contributes visual information derived from the input image to the cross-attention layer, offering visual guidance in the attention mechanism. This prevents $\hat{x}_B$ from strictly adhering to the text instruction, ensuring that the output remains connected to the visual context of $x_A$ and preventing undue deviation.

\begin{figure}[!t]
\begin{center}
\begin{subfigure}[b]{0.32\linewidth}
\includegraphics[width=\linewidth]{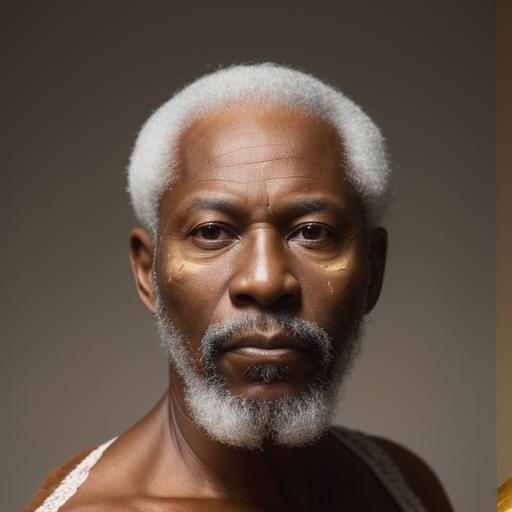} %
\caption{Input}
\end{subfigure}
\begin{subfigure}[b]{0.32\linewidth}
\includegraphics[width=\linewidth]{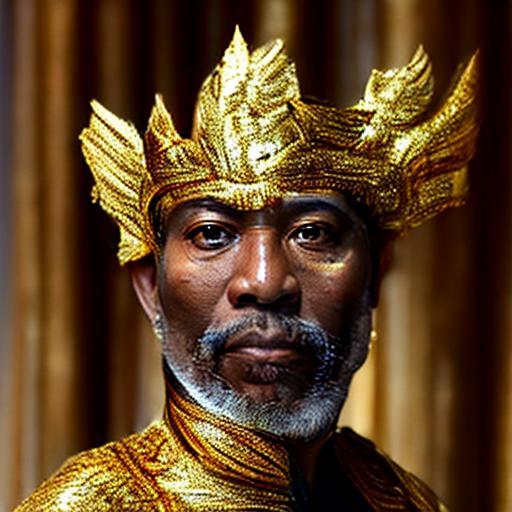} %
\caption{Ours w/o Prt}
\end{subfigure}
\begin{subfigure}[b]{0.32\linewidth}
\includegraphics[width=\linewidth]{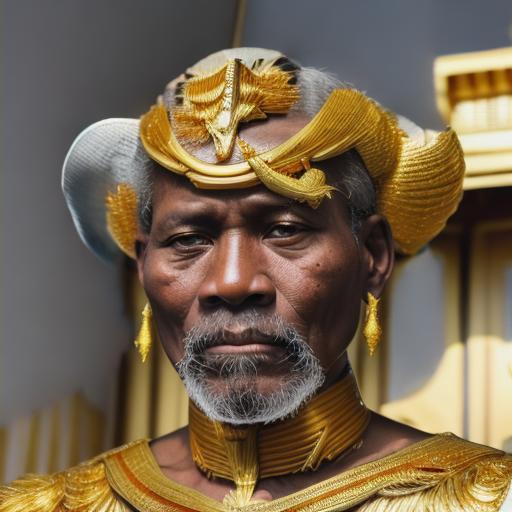} %
\caption{Ours w/o Spt}
\end{subfigure}
\\
\begin{subfigure}[b]{0.32\linewidth}
\includegraphics[width=\linewidth]{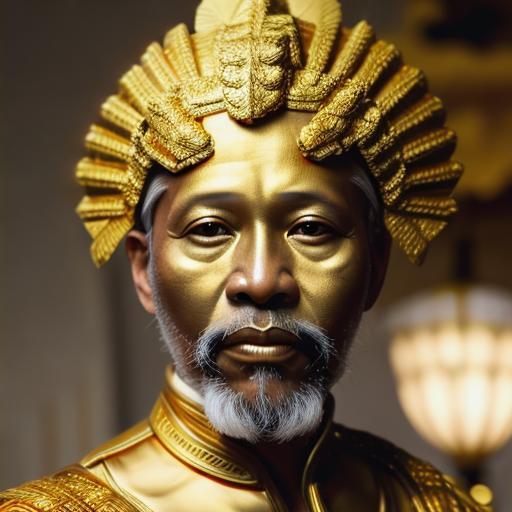} %
\caption{Ours w/o Iemb}
\end{subfigure}
\begin{subfigure}[b]{0.32\linewidth}
\includegraphics[width=\linewidth]{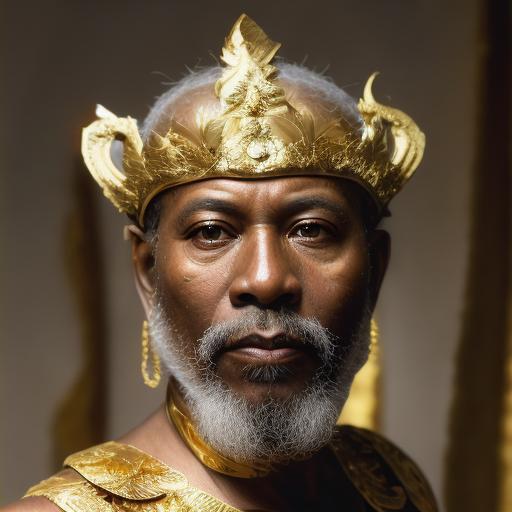} %
\caption{Ours w/o CFG}
\end{subfigure}
\begin{subfigure}[b]{0.32\linewidth}
\includegraphics[width=\linewidth]{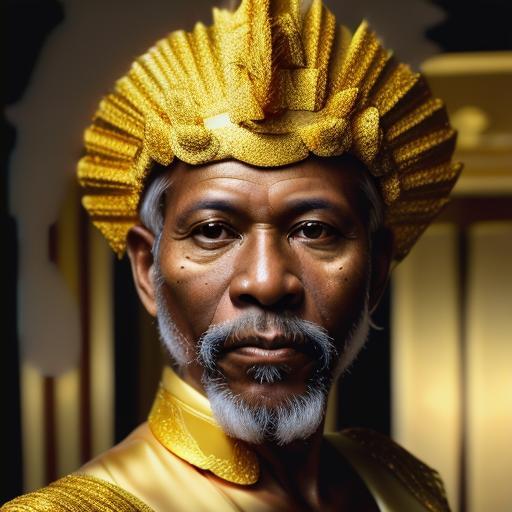} %
\caption{Ours}
\end{subfigure}
\caption{Ablation study of design choice of MCDM, where the goal is to have the person in (a) wear a royal costume.   Training from scratch (b) yields the poorest image quality due to the absence of image generation priors and text prompt interpretation. Dropping spatial embeddings (c) fails to preserve spatial layout and the person's hairstyle. Excluding image embeddings (d) causes "over-editing" towards the target domain, compromising fidelity (\eg, the golden face in (d)). Without classifier-free guidance, less expressive edits emerge (e) (\eg, incomplete crown). In contrast, our full pipeline (f) produces the best editing results.
}
 \ifdraft
\vspace{-3mm}
 \else
\fi
\label{fig:costume_cond_ablation}
\end{center}
\end{figure}


We initializes network weights with pretrained Stable Diffusion~\cite{rombach2021highresolution}. 
The training scheme is similar to Stable Diffusion, but with several differences:
(1) we replace $c_{p_b}$ with $c_{p_a}$ and $x_B$ with $x_A$ by $5 \%$ of time. This  enables the model to reconstruct input images (\ie, perform identical editing), which will be utilized during the inference phase for mask generation. 
(2) Inspired by \cite{karras2023dreampose}, we implement a dropout mechanism for multiple signals for classifier-free guidance. Specifically, with a $20\%$ probability, we drop any combination of the following: $c_s$, $c_{im}$, $c_p$, or even all of them. 

Figure \ref{fig:costume_cond_ablation} illustrates the ablation of these design choices, underscoring the effectiveness of employing all conditional signals simultaneously, as previously discussed.


\begin{figure}[!t]
\begin{center}
\begin{subfigure}[b]{0.32\linewidth}
\includegraphics[width=\linewidth]{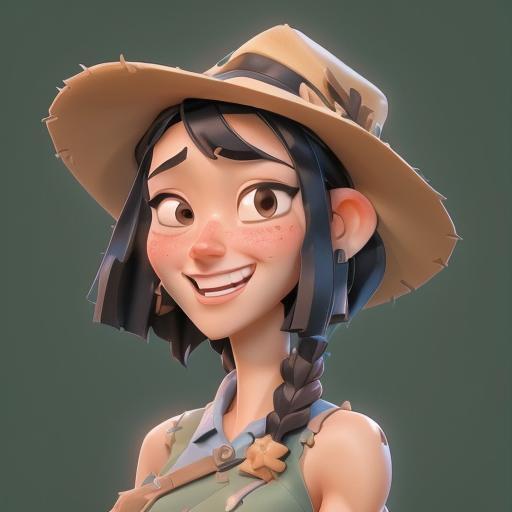} %
\caption{Input}
\end{subfigure}
\begin{subfigure}[b]{0.32\linewidth}
\includegraphics[width=\linewidth]{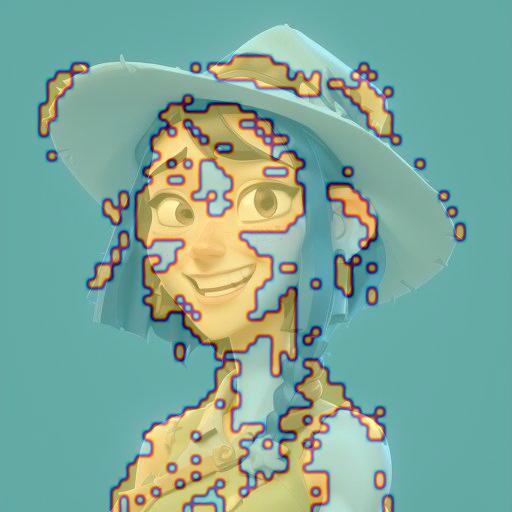} %
\caption{Mask (DiffEdit)}
\end{subfigure}
\begin{subfigure}[b]{0.32\linewidth}
\includegraphics[width=\linewidth]{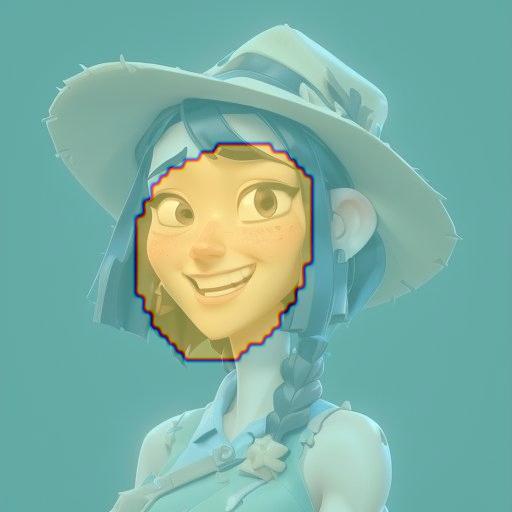} %
\caption{Mask (Our Model)}
\end{subfigure}
\\
\begin{subfigure}[b]{0.32\linewidth}
\includegraphics[width=\linewidth]{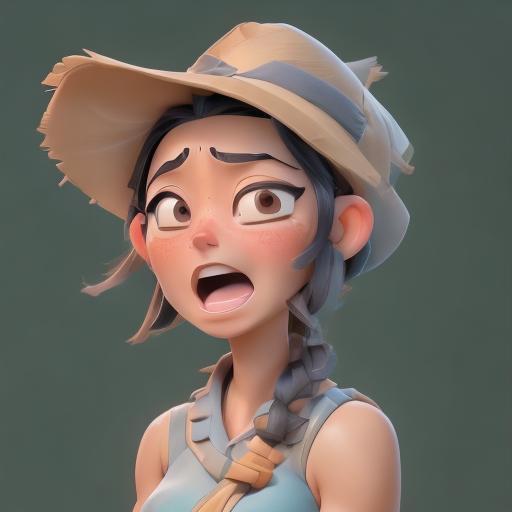} %
\caption{Generate w/o Mask}
\end{subfigure}
\begin{subfigure}[b]{0.32\linewidth}
\includegraphics[width=\linewidth]{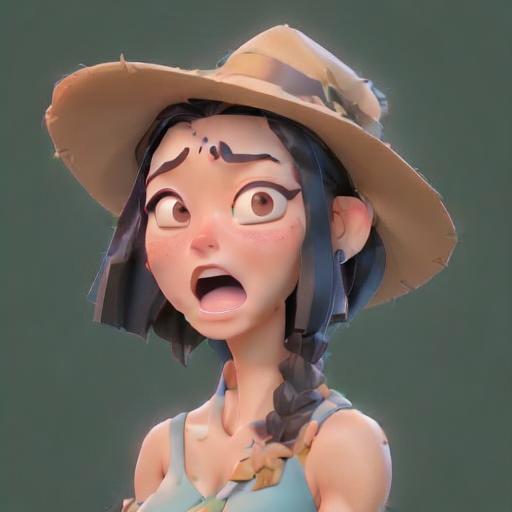} %
\caption{Generate with (b)}
\end{subfigure}
\begin{subfigure}[b]{0.32\linewidth}
\includegraphics[width=\linewidth]{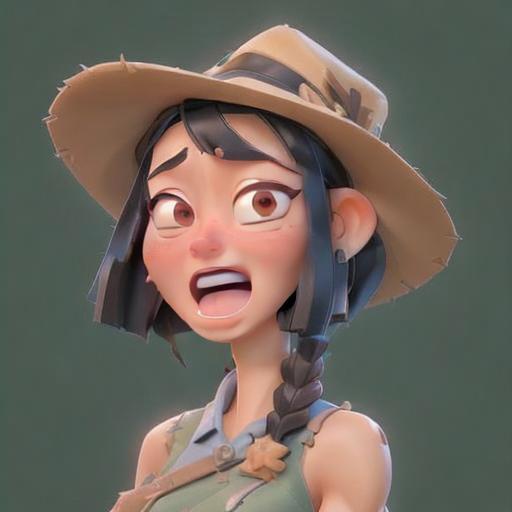} %
\caption{Generate with (c)}
\end{subfigure}
\caption{ Comparison of mask-guided editing on cartoon expression editing (to shocked expression).  Standard generation (d) alters details (\eg, patterns on hats and upper clothing) in the input image (a). Applying the mask generation strategy to our model improves the accuracy of the generated mask (c) compared to the one generated by DiffEdit (b). When guided by the mask in (c), the edited image (f) effectively preserves details (e.g., clothing) compared to the one (see (e)) guided by (b).
}
 \ifdraft
\vspace{-5mm}
 \else
\fi
\label{fig:mask_ablation}
\end{center}
\end{figure}

\subsection{Mask-Guided Editing using Trained Model}

After training, the standard approach for generating predictions $\hat{x}_B$ from $x_A$ involves denoising a random latent $z_T$ over $T$ iterations using trained model (with classifier-free guidance). 
While the generated $\hat{x}_B$ successfully accomplishes the desired edits while preserving identity and layout, challenges may persist in retaining specific details of the subject's features. For example, in Figure \ref{fig:mask_ablation}, an illustration of expression editing (to a shocked expression) depicts the standard generation output (d), where the hat and upper clothing patterns differ from those in the input image (a).  

To enhance the preservation of these details, a mask can be derived from the trained MCDM, providing explicit guidance for the denoising process. This mask indicates areas for editing and those to be left untouched. 
We adapt DiffEdit~\cite{couairon2022diffedit} to automatically generate such a mask. 
The key difference between our and DiffEdit's mask generation strategy is that, instead of relying on a pretrained Stable Diffusion model, we leverage our trained MCDM with its reconstruction capabilities to achieve more precise mask generation.
By applying DiffEdit to our trained MCDM instead of the original Stable Diffusion model, we can achieve more precise mask generation due to MCDM's reconstruction capability.

Figure \ref{fig:mask_ablation} (c) shows an example of editing mask  generated by our trained model, which is more accurate than the one produced by the DiffEditt used to produce pairs (Figure \ref{fig:mask_ablation} (b)). 
This demonstration underscores the MCDM's capacity to discern the types of content that should be edited, even by training on an imperfect dataset.

Once we have the mask $M$, at each timestep $t$, we calculate the mask-guided predicted noise by:
\begin{align}
   \hat{\epsilon} =&  \;   \tilde{\epsilon}_{\theta}(z_t, t,  \{c_s, c_{im}, c_{p_b}\}) \odot  M  \; + \\& \; \tilde{\epsilon}_{\theta}(z_t, t,  \{c_s, c_{im}, c_{p_a}\}) \odot  (1-M).   \nonumber 
\end{align}
It implies that we denoise for target editing (using $p_b$) within the mask, and preserve the original image content (using $p_a$) outside the mask. Figure \ref{fig:mask_ablation} (e) shows the result with mask guidance. 
See implementation details in supplementary.


\section{Experiments}
\label{sec:exp}
\noindent \textbf{Datasets}:
We evaluate the performance of our pipelines in two distinct portrait editing tasks: costume editing and cartoon expression editing. 
For each task, we define four different editing directions for input in a specific domain. 
For costume editing,  the input image is a realistic portrait image with everyday costume, and the output is the same person with flower, sheep, Santa Claus, or royal costume.  
For cartoon expression editing, the input image is a cartoon portrait with a neutral expression, while the output is the same cartoon character with four different expressions: angry, shocked, laughing, or crying.
For each task, we generate a dataset of 69,900 image pairs (17475 for each editing direction) for training. 
The in-the-wild images for testing are from \cite{Rothe-IJCV-2018}.
See details in supplementary.


\noindent \textbf{Baselines}:
We choose 6 state-of-the-art image editing baselines for comparison. In particular,  Prompt2Prompt~\cite{hertz2022prompt}, pix2pix-zero~\cite{parmar2023zero}, DiffEdit~\cite{couairon2022diffedit}, SDEdit~\cite{meng2021sdedit} are training-free diffusion methods  with editing direction guided by text prompt. 
Since SDEdit is sensitive to a strength parameter, we test two different parameters of it, namely SDEdit 0.5 and SDEdit 0.8. Larger strength produces outputs that obeys the editing directions but deviates from the input images. 
SPADE~\cite{park2019SPADE} and BBDM~\cite{li2023bbdm} are training-based image editing framework building on top of Generative Adversarial Networks~\cite{creswell2018generative} and diffusion model~\cite{croitoru2023diffusion}, respectively. 

\begin{figure*}[!t]
\begin{center}
\begin{minipage}{0.12\linewidth}
\vspace{-60pt}
\begin{tabular}{c}
\quad Flower
\end{tabular}
\end{minipage}
\begin{subfigure}[b]{0.133\linewidth}
\includegraphics[width=\linewidth]{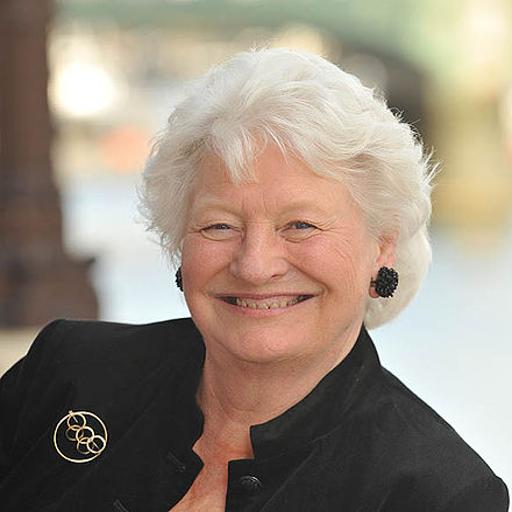} %
\end{subfigure}
\begin{subfigure}[b]{0.133\linewidth}
\includegraphics[width=\linewidth]{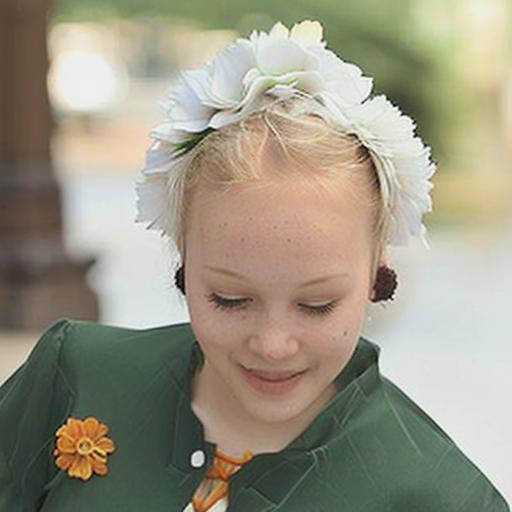} %
\end{subfigure}
\begin{subfigure}[b]{0.133\linewidth}
\includegraphics[width=\linewidth]{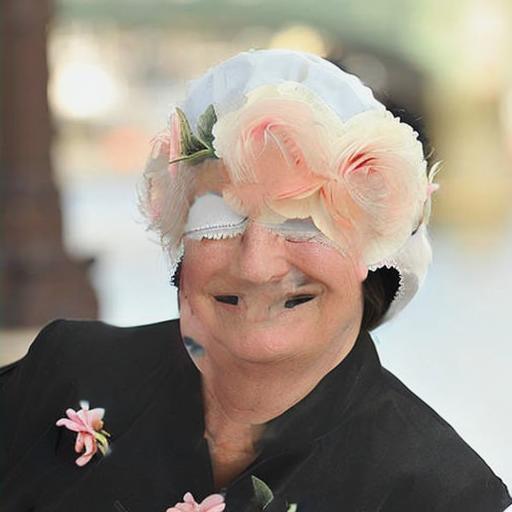} %
\end{subfigure}
\begin{subfigure}[b]{0.133\linewidth}
\includegraphics[width=\linewidth]{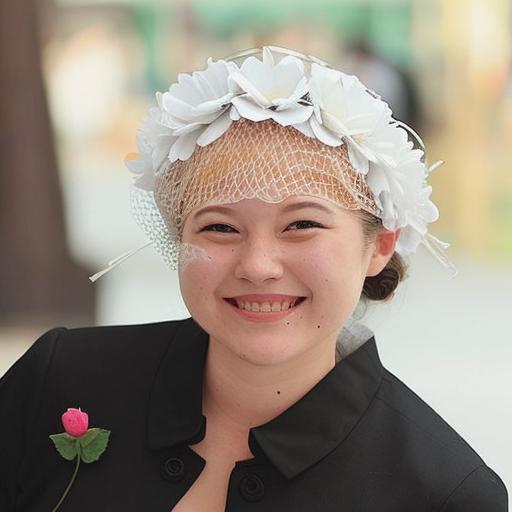} %
\end{subfigure}
\begin{subfigure}[b]{0.133\linewidth}
\includegraphics[width=\linewidth]{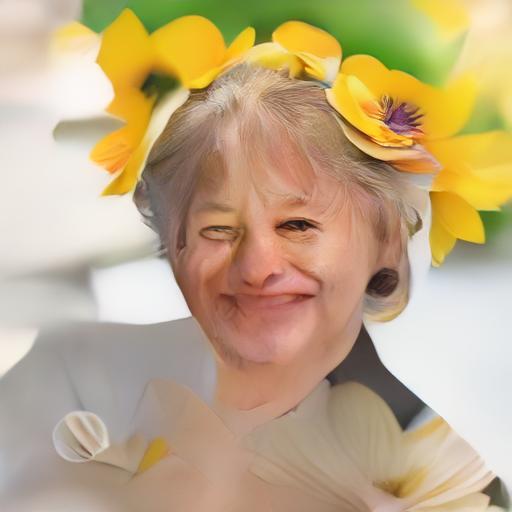} %
\end{subfigure}
\begin{subfigure}[b]{0.133\linewidth}
\includegraphics[width=\linewidth]{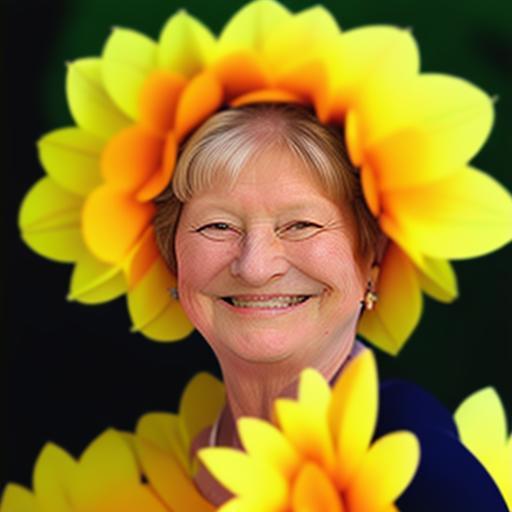} %
\end{subfigure}
\\
\begin{minipage}{0.12\linewidth}
\vspace{-60pt}
\begin{tabular}{c}
\quad Sheep 
\end{tabular}
\end{minipage}
\begin{subfigure}[b]{0.133\linewidth}
\includegraphics[width=\linewidth]{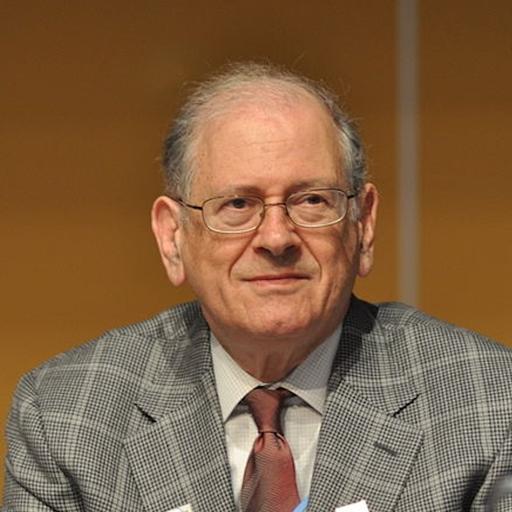} %
\end{subfigure}
\begin{subfigure}[b]{0.133\linewidth}
\includegraphics[width=\linewidth]{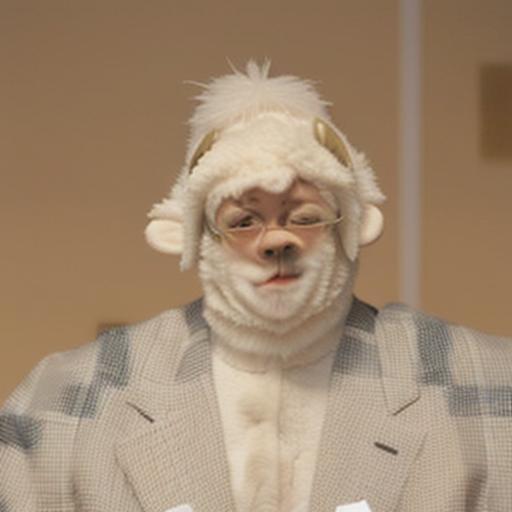} %
\end{subfigure}
\begin{subfigure}[b]{0.133\linewidth}
\includegraphics[width=\linewidth]{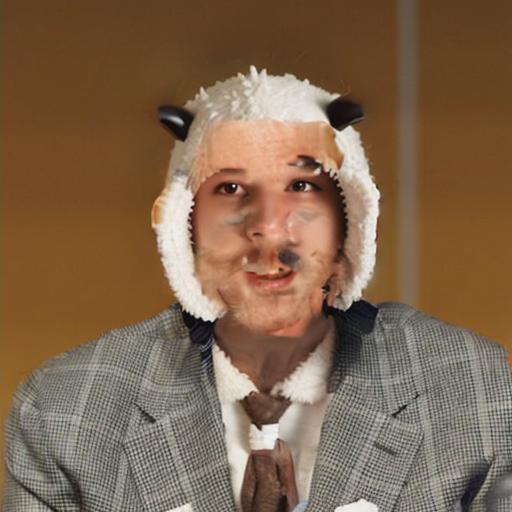} %
\end{subfigure}
\begin{subfigure}[b]{0.133\linewidth}
\includegraphics[width=\linewidth]{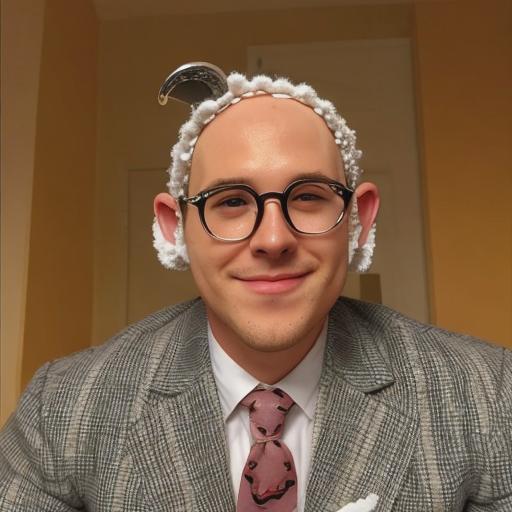} %
\end{subfigure}
\begin{subfigure}[b]{0.133\linewidth}
\includegraphics[width=\linewidth]{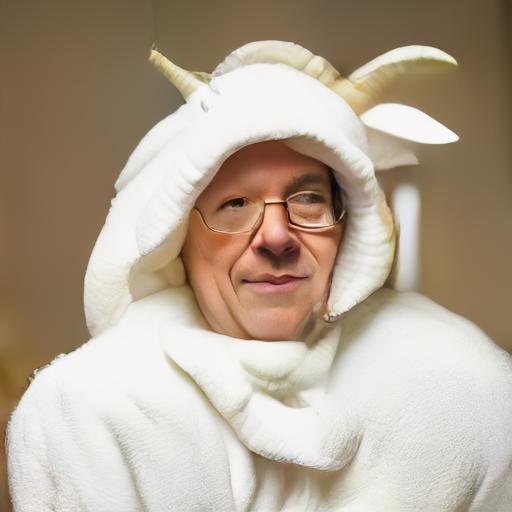} %
\end{subfigure}
\begin{subfigure}[b]{0.133\linewidth}
\includegraphics[width=\linewidth]{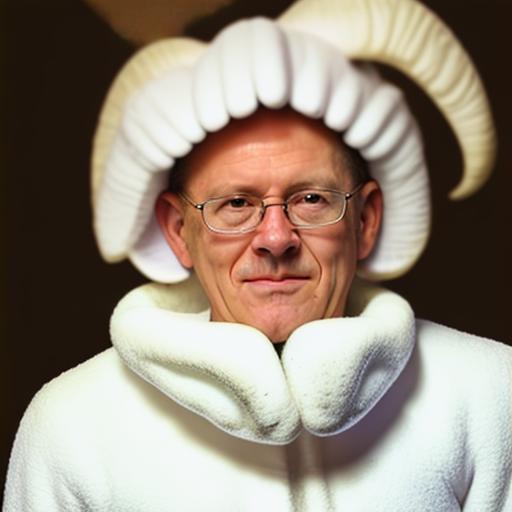} %
\end{subfigure}
\\
\begin{minipage}{0.12\linewidth}
\vspace{-60pt}
\begin{tabular}{c}
\quad Santa \\ \quad Claus
\end{tabular}
\end{minipage}
\begin{subfigure}[b]{0.133\linewidth}
\includegraphics[width=\linewidth]{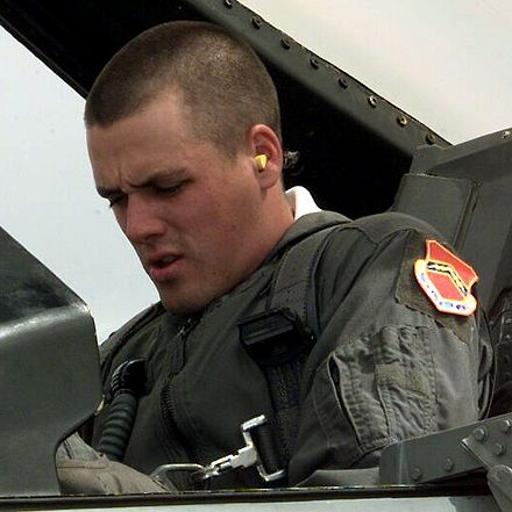} %
\end{subfigure}
\begin{subfigure}[b]{0.133\linewidth}
\includegraphics[width=\linewidth]{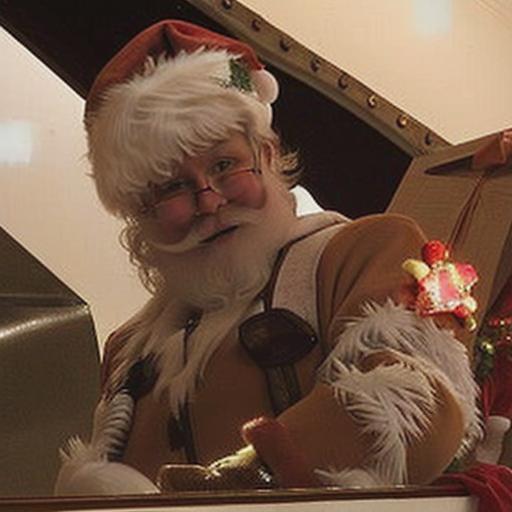} %
\end{subfigure}
\begin{subfigure}[b]{0.133\linewidth}
\includegraphics[width=\linewidth]{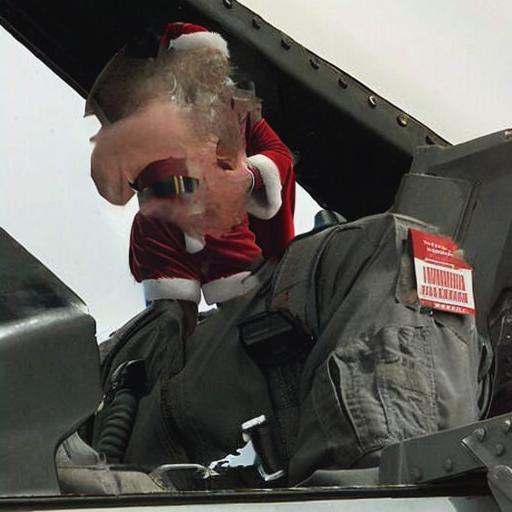} %
\end{subfigure}
\begin{subfigure}[b]{0.133\linewidth}
\includegraphics[width=\linewidth]{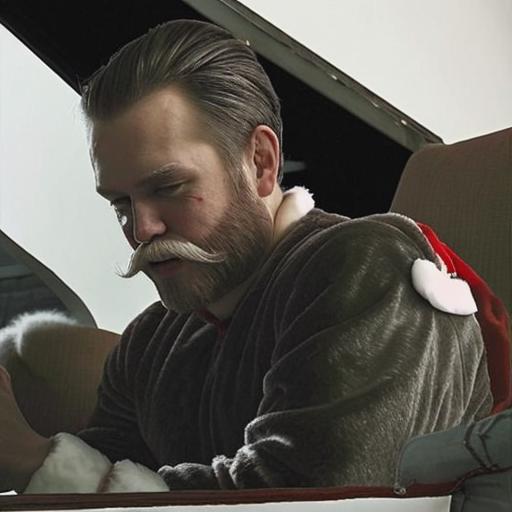} %
\end{subfigure}
\begin{subfigure}[b]{0.133\linewidth}
\includegraphics[width=\linewidth]{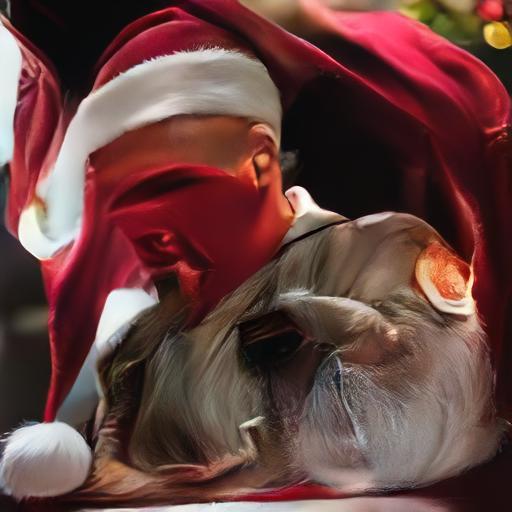} %
\end{subfigure}
\begin{subfigure}[b]{0.133\linewidth}
\includegraphics[width=\linewidth]{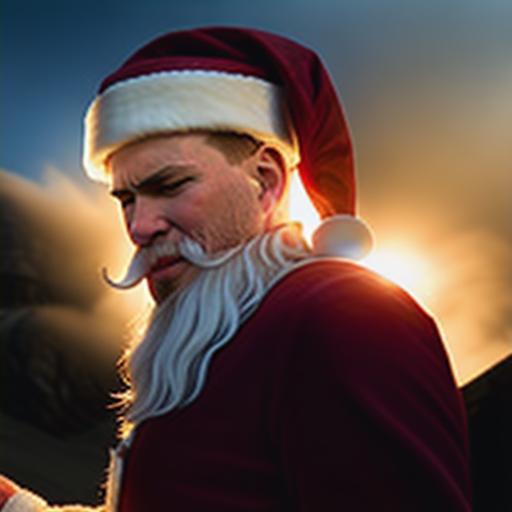} %
\end{subfigure}
\\
\begin{minipage}{0.12\linewidth}
\vspace{-60pt}
\begin{tabular}{c}
\quad Royal 
\end{tabular}
\end{minipage}
\begin{subfigure}[b]{0.133\linewidth}
\includegraphics[width=\linewidth]{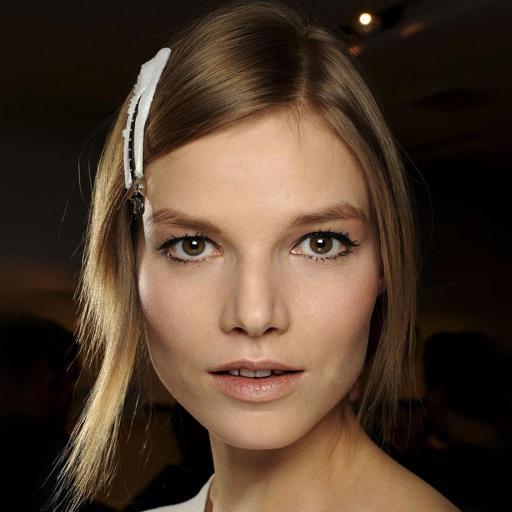} %
\end{subfigure}
\begin{subfigure}[b]{0.133\linewidth}
\includegraphics[width=\linewidth]{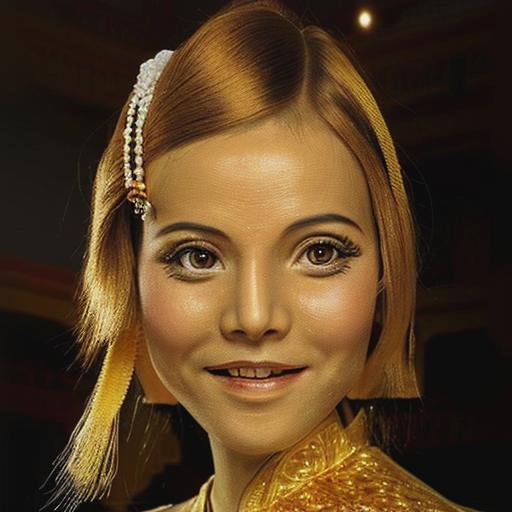} %
\end{subfigure}
\begin{subfigure}[b]{0.133\linewidth}
\includegraphics[width=\linewidth]{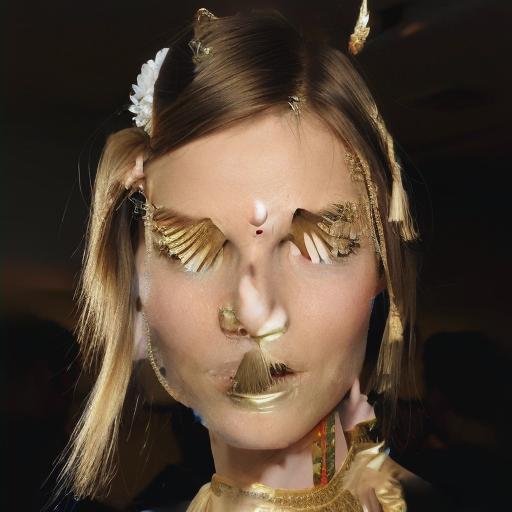} %
\end{subfigure}
\begin{subfigure}[b]{0.133\linewidth}
\includegraphics[width=\linewidth]{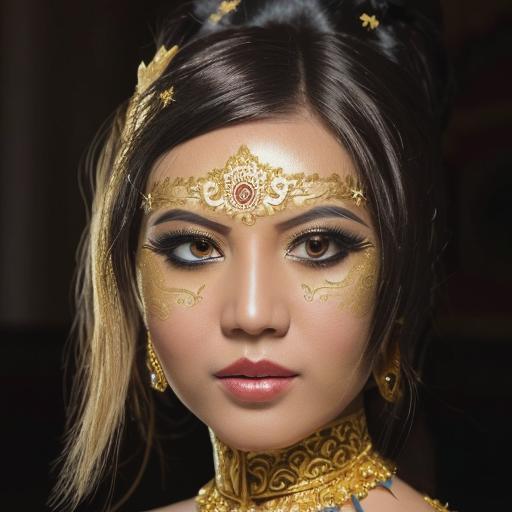} %
\end{subfigure}
\begin{subfigure}[b]{0.133\linewidth}
\includegraphics[width=\linewidth]{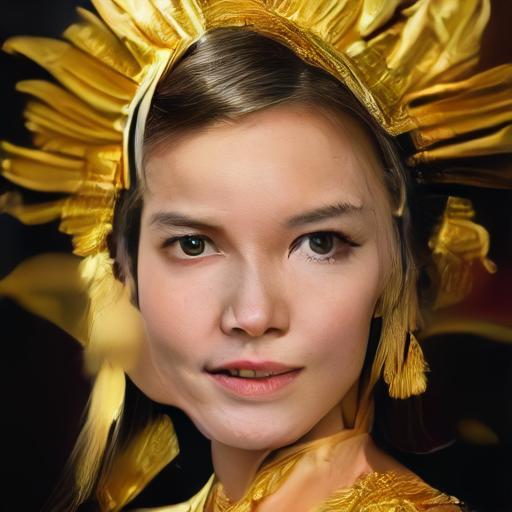} %
\end{subfigure}
\begin{subfigure}[b]{0.133\linewidth}
\includegraphics[width=\linewidth]{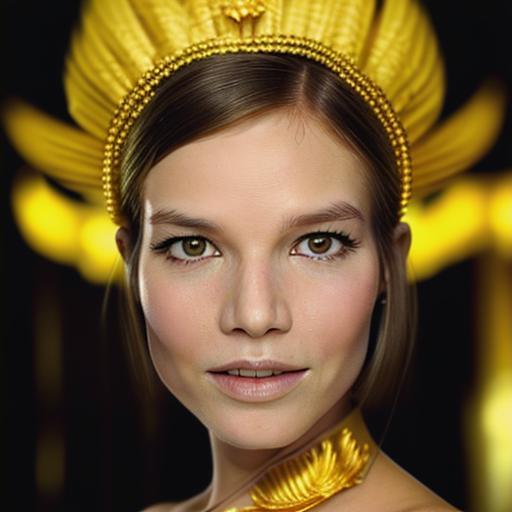} %
\end{subfigure}
\\
\begin{minipage}{0.12\linewidth}
\vspace{-60pt}
\begin{tabular}{c}
\quad Angry
\end{tabular}
\end{minipage}
\begin{subfigure}[b]{0.133\linewidth}
\includegraphics[width=\linewidth]{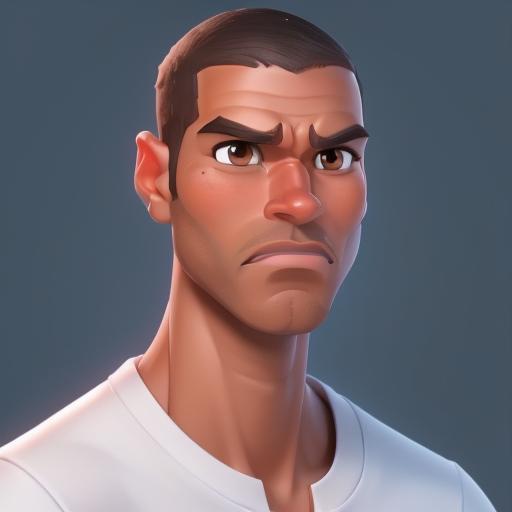} %
\end{subfigure}
\begin{subfigure}[b]{0.133\linewidth}
\includegraphics[width=\linewidth]{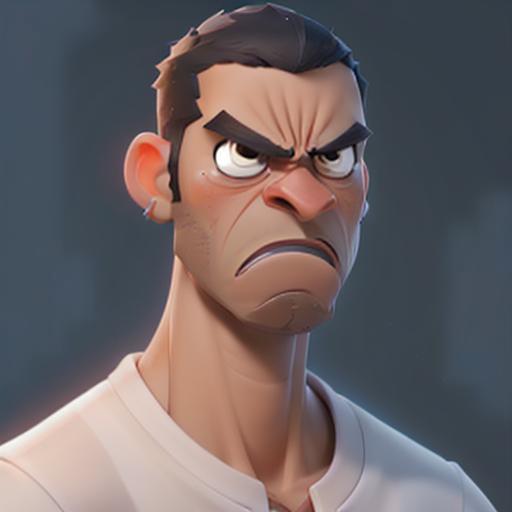} %
\end{subfigure}
\begin{subfigure}[b]{0.133\linewidth}
\includegraphics[width=\linewidth]{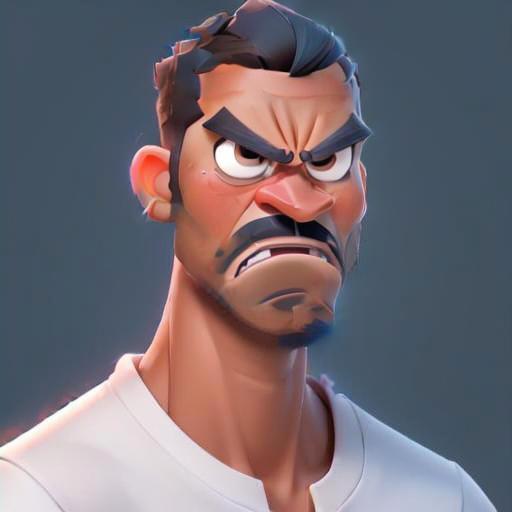} %
\end{subfigure}
\begin{subfigure}[b]{0.133\linewidth}
\includegraphics[width=\linewidth]{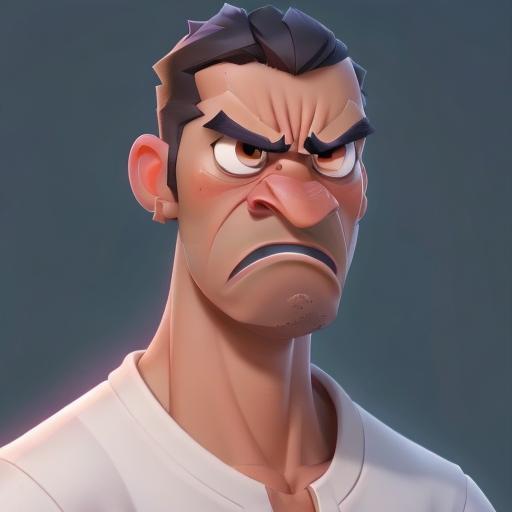} %
\end{subfigure}
\begin{subfigure}[b]{0.133\linewidth}
\includegraphics[width=\linewidth]{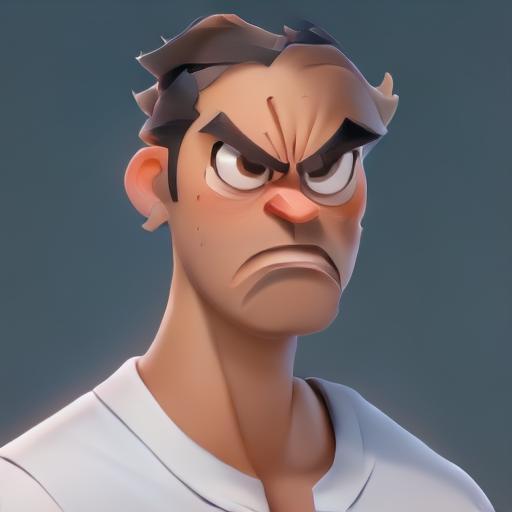} %
\end{subfigure}
\begin{subfigure}[b]{0.133\linewidth}
\includegraphics[width=\linewidth]{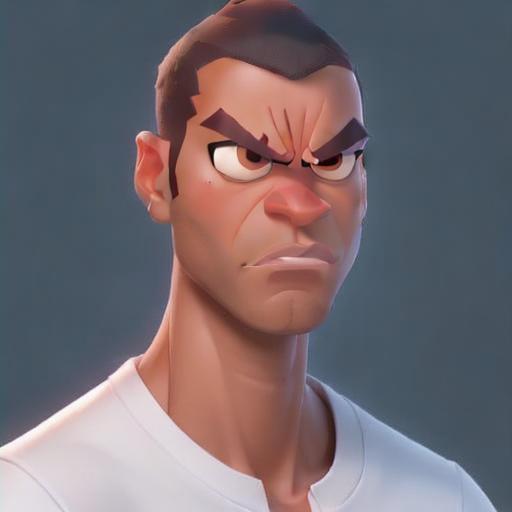} %
\end{subfigure}
\\
\begin{minipage}{0.12\linewidth}
\vspace{-60pt}
\begin{tabular}{c}
\quad Shocked
\end{tabular}
\end{minipage}
\begin{subfigure}[b]{0.133\linewidth}
\includegraphics[width=\linewidth]{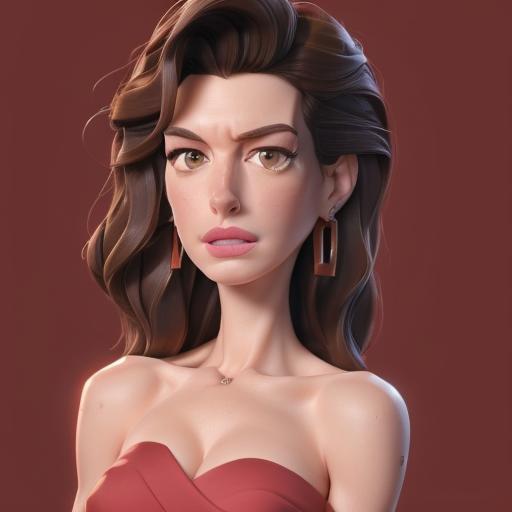} %
\end{subfigure}
\begin{subfigure}[b]{0.133\linewidth}
\includegraphics[width=\linewidth]{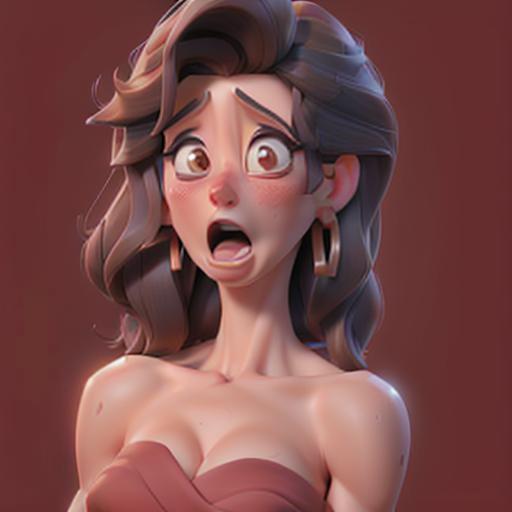} %
\end{subfigure}
\begin{subfigure}[b]{0.133\linewidth}
\includegraphics[width=\linewidth]{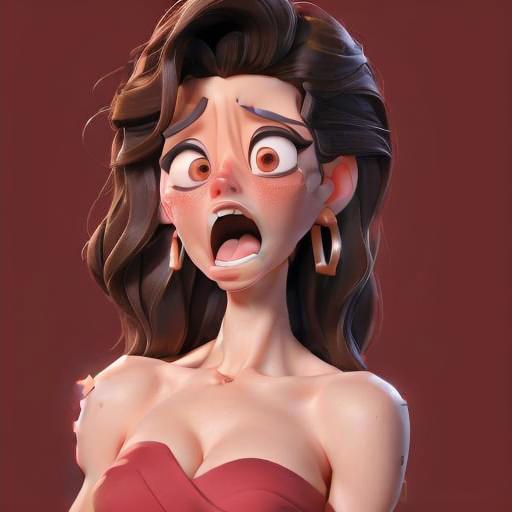} %
\end{subfigure}
\begin{subfigure}[b]{0.133\linewidth}
\includegraphics[width=\linewidth]{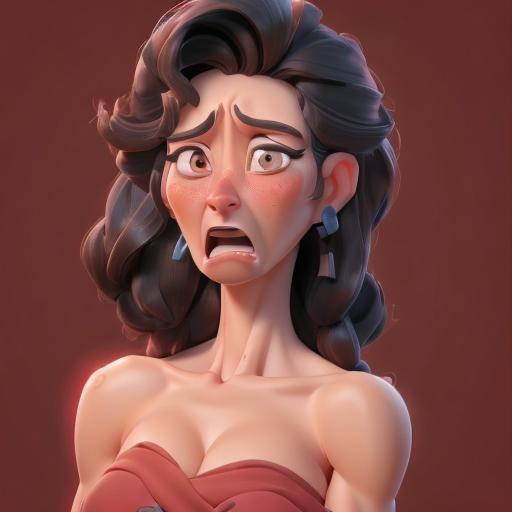} %
\end{subfigure}
\begin{subfigure}[b]{0.133\linewidth}
\includegraphics[width=\linewidth]{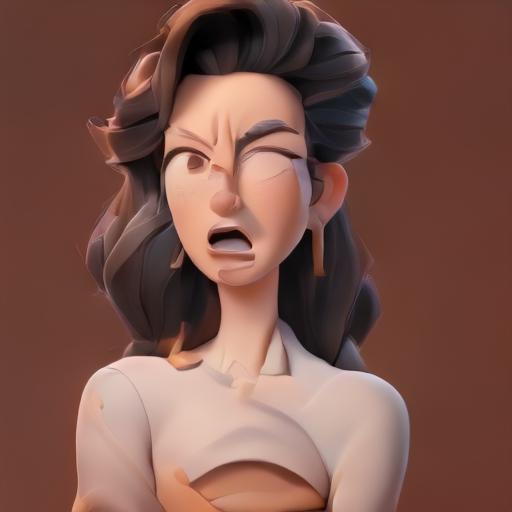} %
\end{subfigure}
\begin{subfigure}[b]{0.133\linewidth}
\includegraphics[width=\linewidth]{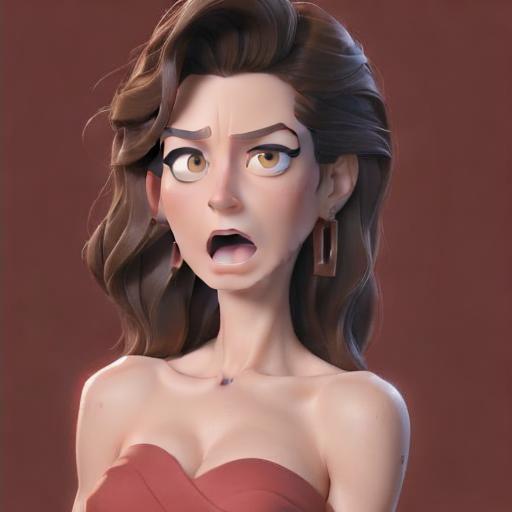} %
\end{subfigure}
\\
\begin{minipage}{0.12\linewidth}
\vspace{-60pt}
\begin{tabular}{c}
\quad Laughing
\end{tabular}
\end{minipage}
\begin{subfigure}[b]{0.133\linewidth}
\includegraphics[width=\linewidth]{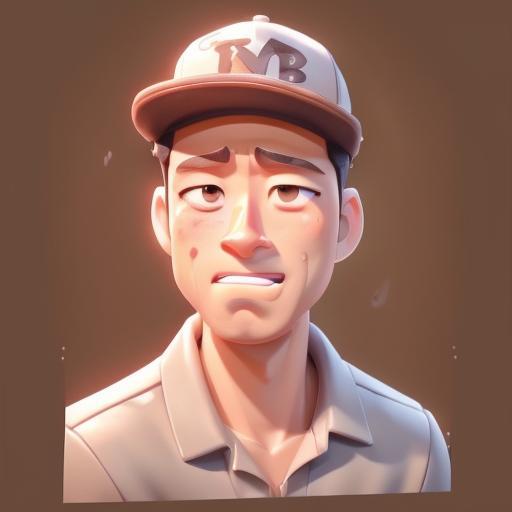} %
\end{subfigure}
\begin{subfigure}[b]{0.133\linewidth}
\includegraphics[width=\linewidth]{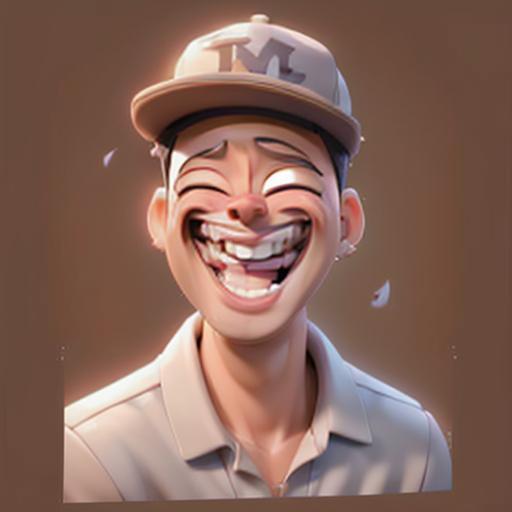} %
\end{subfigure}
\begin{subfigure}[b]{0.133\linewidth}
\includegraphics[width=\linewidth]{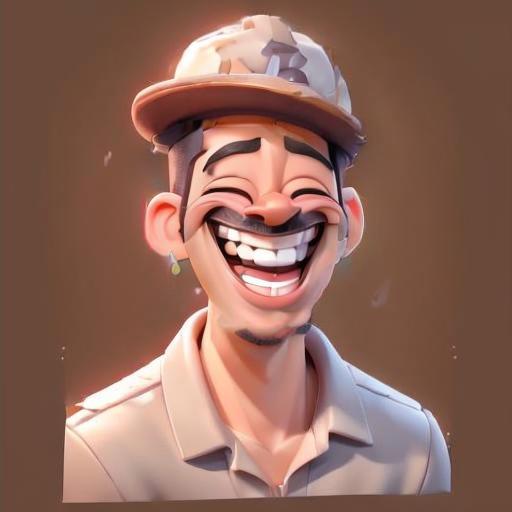} %
\end{subfigure}
\begin{subfigure}[b]{0.133\linewidth}
\includegraphics[width=\linewidth]{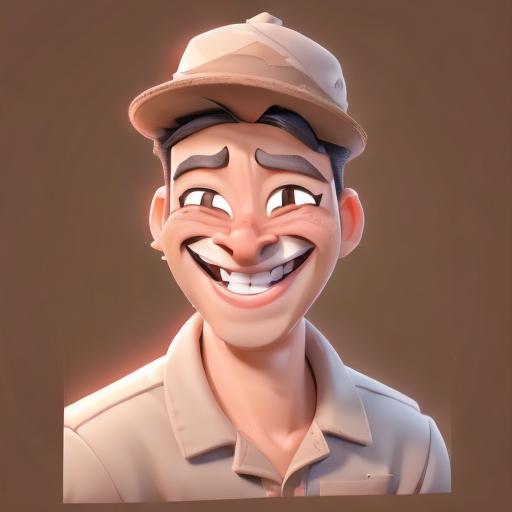} %
\end{subfigure}
\begin{subfigure}[b]{0.133\linewidth}
\includegraphics[width=\linewidth]{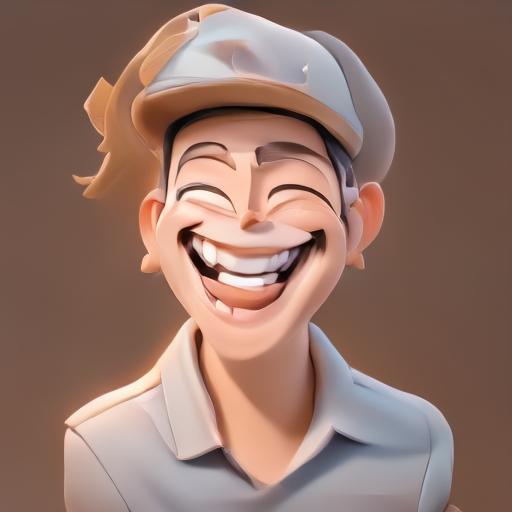} %
\end{subfigure}
\begin{subfigure}[b]{0.133\linewidth}
\includegraphics[width=\linewidth]{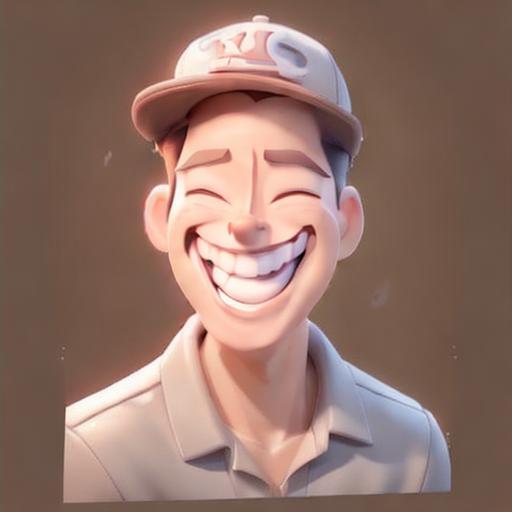} %
\end{subfigure}
\\
\begin{minipage}{0.12\linewidth}
\vspace{-87pt}
\begin{tabular}{c}
\quad Crying
\end{tabular}
\end{minipage}
\begin{subfigure}[b]{0.133\linewidth}
\includegraphics[width=\linewidth]{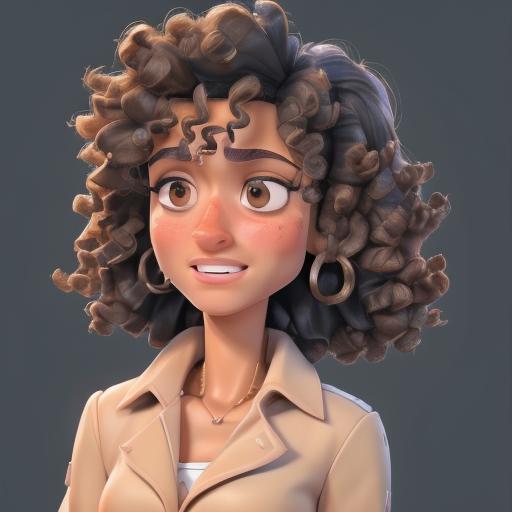} %
\caption{Input}
\end{subfigure}
\begin{subfigure}[b]{0.133\linewidth}
\includegraphics[width=\linewidth]{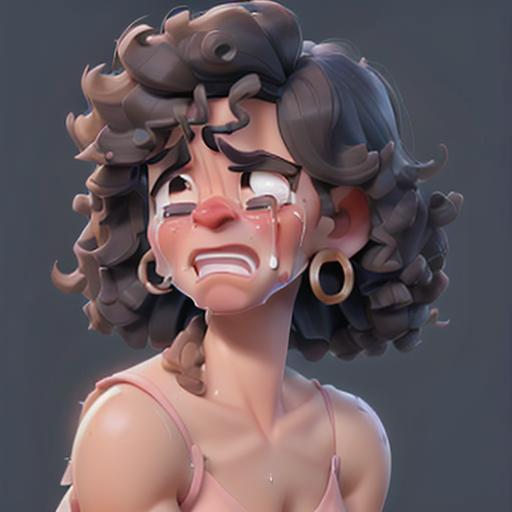} %
\caption{Prompt2Prompt}
\end{subfigure}
\begin{subfigure}[b]{0.133\linewidth}
\includegraphics[width=\linewidth]{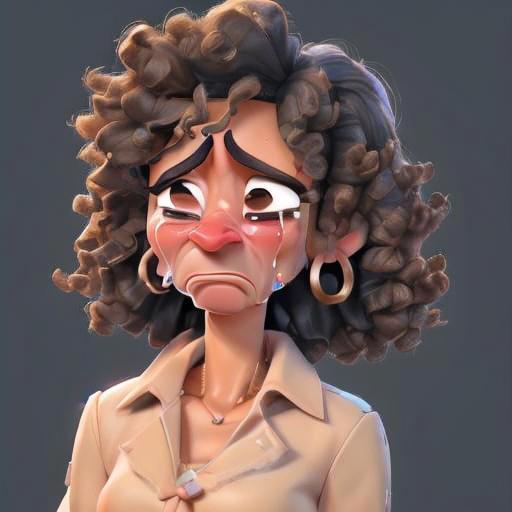} %
\caption{DiffEdit}
\end{subfigure}
\begin{subfigure}[b]{0.133\linewidth}
\includegraphics[width=\linewidth]{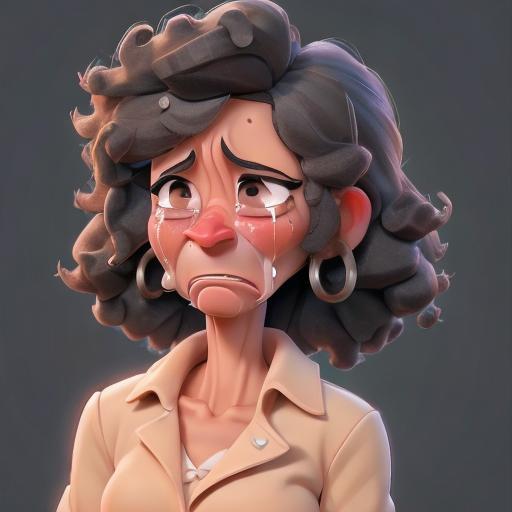} %
\caption{SDEdit 0.5}
\end{subfigure}
\begin{subfigure}[b]{0.133\linewidth}
\includegraphics[width=\linewidth]{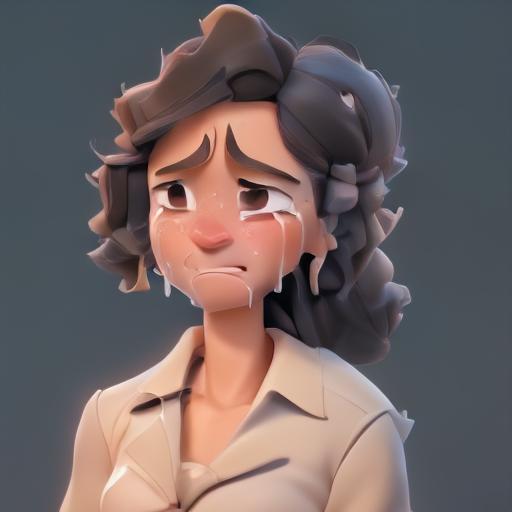} %
\caption{BBDM}
\end{subfigure}
\begin{subfigure}[b]{0.133\linewidth}
\includegraphics[width=\linewidth]{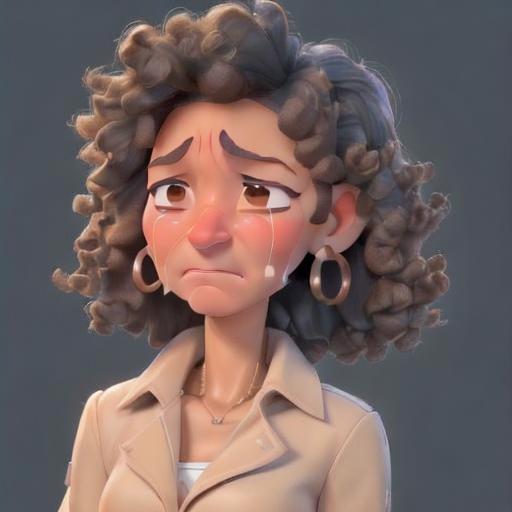} %
\caption{Ours}
\end{subfigure}
\vspace{-2mm}
\caption{
In-the-Wild results comparison: existing methods either fail to apply desired edits (\eg, SDEdit in first 4 rows) or struggle to figure out which region to apply the edits (\eg, DiffEdit in first 4 rows). When the edits do take effect, they alter input features too much that destroy the subject's identity (\eg, facial hair in row 5 (c), arm muscle in row 6 (d)), or create significant artifacts (\eg, Prompt2Prompt and BBDM). By contrast, our model can preserve input subjects' appearance features well and achieve desired editing at high visual quality.
}
\vspace{-2mm}
\label{fig:baseline_comparison_costume}
\end{center}
\end{figure*}

\begin{table}
\resizebox{0.48\textwidth}{!}{
\begin{tabular}{|l|lll|lll|}
 \hline
& \multicolumn{3}{c|}{Costume Editing Dataset} & \multicolumn{3}{c|}{Expression Editing Dataset} \\ \hline
& SSIM & LPIPS & FID & SSIM & LPIPS & FID \\ \hline
Prompt2Prompt & 0.764 & 0.694 & 83.94 & 0.936 & 0.426 & 33.13 \\
pix2pix-zero & 0.695 & 0.812 & 120.8 & 0.897 & 0.517 & 46.65 \\
DiffEdit & 0.722 & 0.721 & 93.52 & 0.922 & 0.423 & 32.20 \\
SDEdit 0.8 & 0.681 & 0.766 & 93.17 & 0.914 & 0.461 & 38.59 \\
SDEdit 0.5 & 0.713 & 0.712 & 68.85 & 0.925 & 0.445 & 38.99 \\
SPADE & 0.761 & 0.651 & 67.96 & 0.936 & 0.427 & 31.02 \\
BBDM & 0.776 & 0.676 & 64.17 & 0.937 & 0.432 & 27.26 \\ \hline
Ours w/o Prt & 0.734 & 0.672 & 84.58 & 0.940 & 0.420 & 28.82 \\
Ours w/o Spt & 0.767 & 0.665 & 78.38 & 0.735 & 0.430 & 24.56 \\
Ours w/o Iemb & 0.785 & 0.636 & 65.74 & 0.941 & 0.408 & 23.07 \\
Ours w/o Mask & 0.788 & 0.639 & 63.52 & 0.940 & 0.421 & 26.61 \\ \hline
Ours & \textbf{0.791} & \textbf{0.633} & \textbf{52.56} & \textbf{0.943} & \textbf{0.400} & \textbf{22.09} \\ \hline
\end{tabular}
}
\caption{Quantitative results of all tested methods, where our method outperforms all tested baselines and variants over all metrics. Please note, there are no metrics that can accurately describe the performance of models because the ground truths we used are not real and unique truths.}
\label{table:baseline_comparison}
\end{table}

\begin{figure*}[!h]
\begin{center}
\begin{subfigure}[b]{0.135\linewidth}
\includegraphics[width=\linewidth]{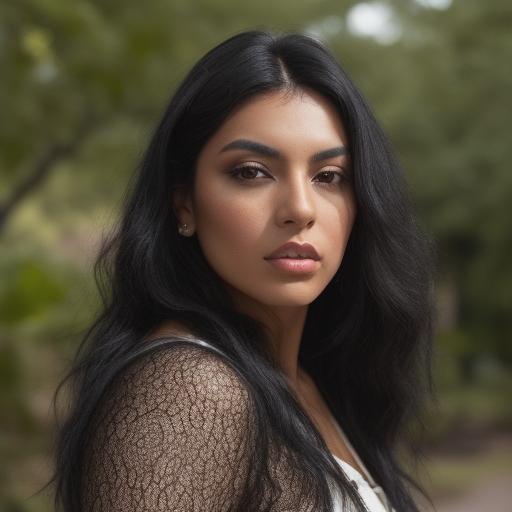} %
\end{subfigure}
\begin{subfigure}[b]{0.135\linewidth}
\includegraphics[width=\linewidth]{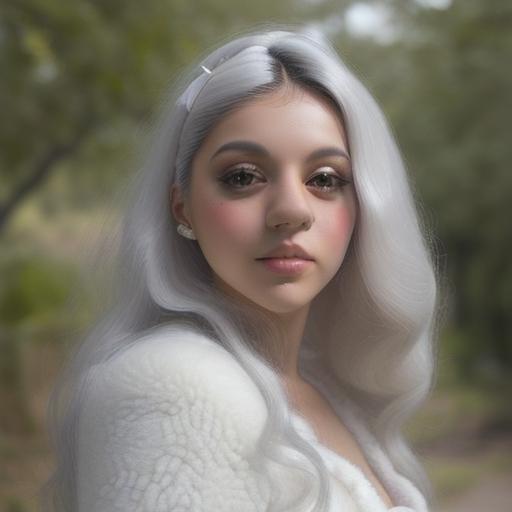} %
\end{subfigure}
\begin{subfigure}[b]{0.135\linewidth}
\includegraphics[width=\linewidth]{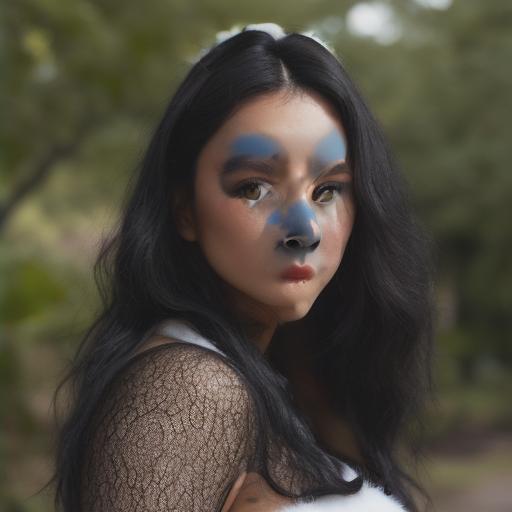} %
\end{subfigure}
\begin{subfigure}[b]{0.135\linewidth}
\includegraphics[width=\linewidth]{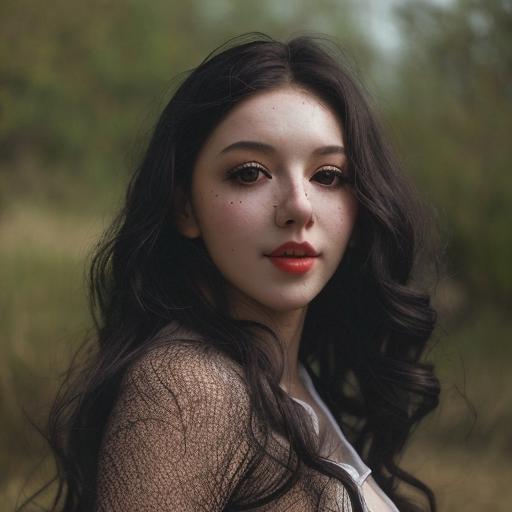} %
\end{subfigure}
\begin{subfigure}[b]{0.135\linewidth}
\includegraphics[width=\linewidth]{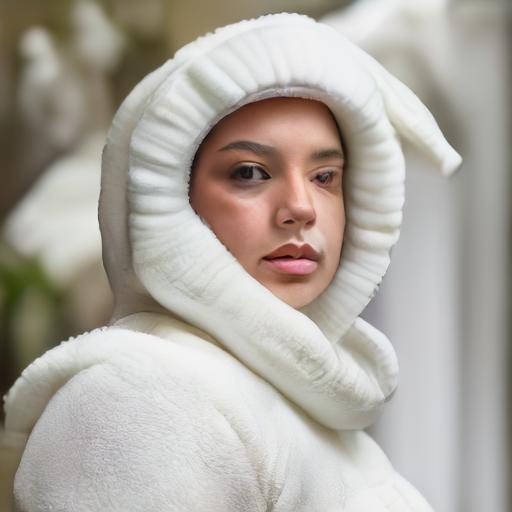} %
\end{subfigure}
\begin{subfigure}[b]{0.135\linewidth}
\includegraphics[width=\linewidth]{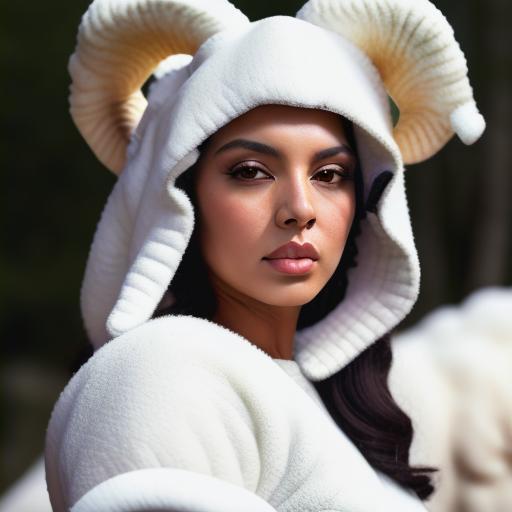} %
\end{subfigure}
\begin{subfigure}[b]{0.135\linewidth}
\includegraphics[width=\linewidth]{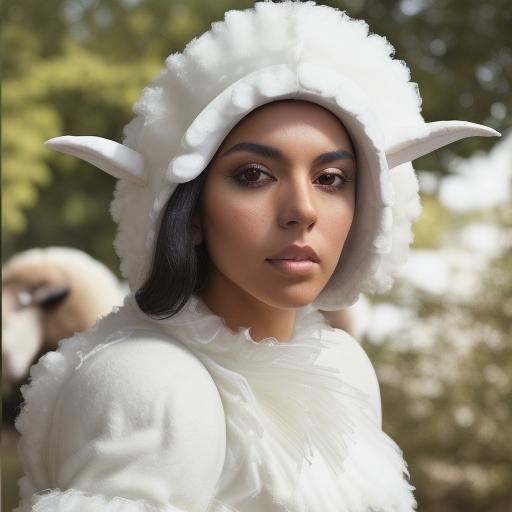} %
\end{subfigure}
\\
\begin{subfigure}[b]{0.135\linewidth}
\includegraphics[width=\linewidth]{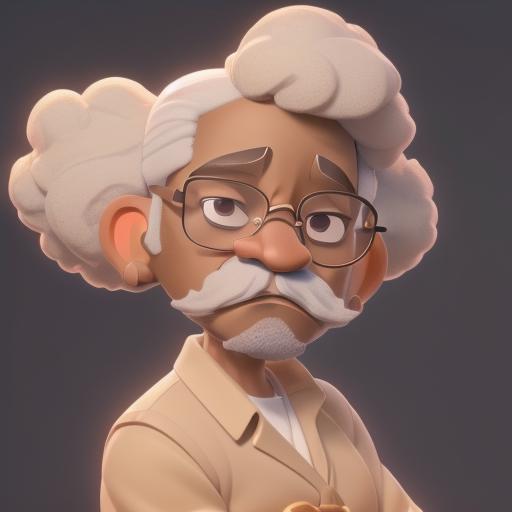} %
\caption{Input}
\end{subfigure}
\begin{subfigure}[b]{0.135\linewidth}
\includegraphics[width=\linewidth]{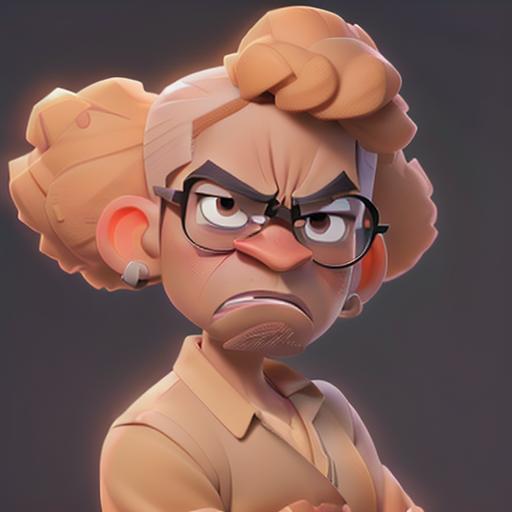} %
\caption{Prompt2Prompt}
\end{subfigure}
\begin{subfigure}[b]{0.135\linewidth}
\includegraphics[width=\linewidth]{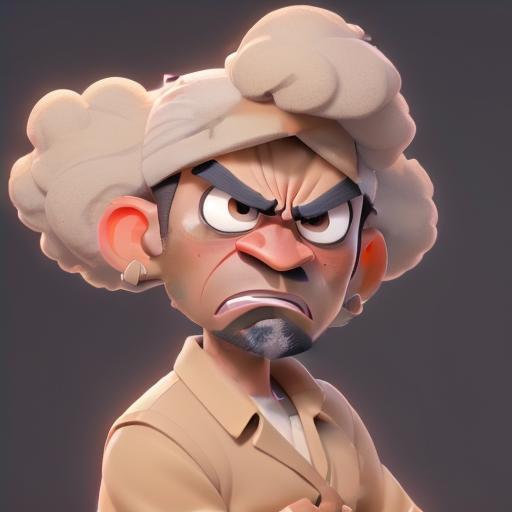} %
\caption{DiffEdit}
\end{subfigure}
\begin{subfigure}[b]{0.135\linewidth}
\includegraphics[width=\linewidth]{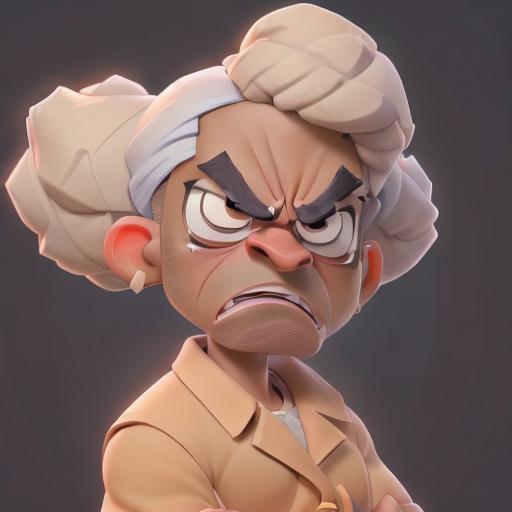} %
\caption{SDEdit 0.5}
\end{subfigure}
\begin{subfigure}[b]{0.135\linewidth}
\includegraphics[width=\linewidth]{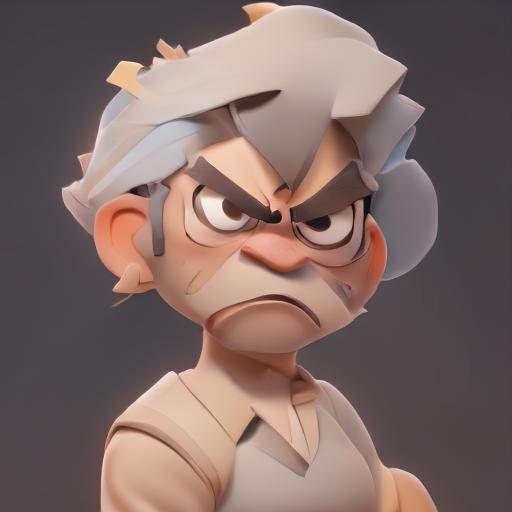} %
\caption{BBDM}
\end{subfigure}
\begin{subfigure}[b]{0.135\linewidth}
\includegraphics[width=\linewidth]{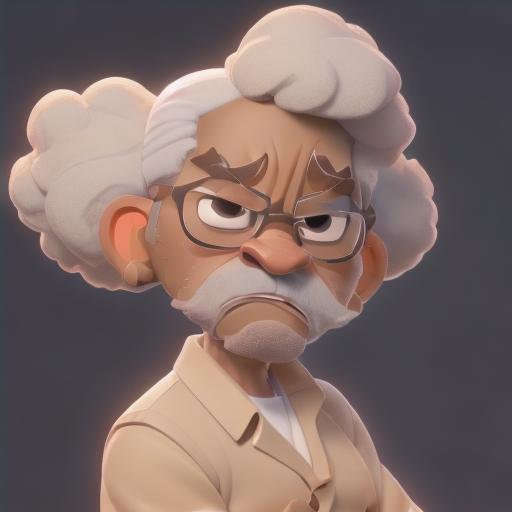} %
\caption{Ours}
\end{subfigure}
\begin{subfigure}[b]{0.135\linewidth}
\includegraphics[width=\linewidth]{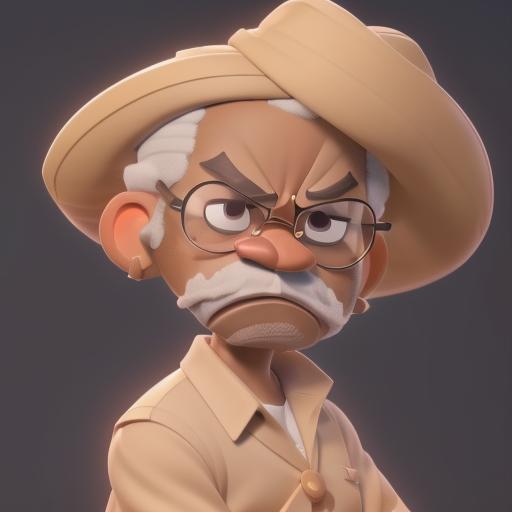} %
\caption{Generated GT}
\end{subfigure}
\caption{Comparison on validation set. In row 1 (sheep costume), the training-free baselines (b) to (d) fall short of achieving the intended edits , while the training-based methods (e) exhibit noticeable artifacts on eyes.
In row 2 (angry), all baselines change the subject identity (\eg, missing glasses, wrong hair color and clothing). In contrast, our method produces high-quality editing results while preserving the identity. Note that validation set input is from generated pairs, so baseline results look better than Figure~\ref{fig:baseline_comparison_costume}.
}
 \ifdraft
\vspace{-6mm}
 \else
\fi
\label{fig:val_baseline_comparison_expression}
\end{center}

\end{figure*}

\noindent \textbf{Real-World Applications}:
We showcase the practical applications of models trained on two datasets through two distinct scenarios.
The first application revolves around real portrait costume editing, wherein the inputs are in-the-wild portrait images. As shown in top 4 rows in Figure~\ref{fig:baseline_comparison_costume}, both training-based and training-free methods yield unsatisfactory results; the former exhibits noticeable artifacts, while the latter often fails to align with the provided prompts.

The second application is sticker pack generation. Here, the objective is to generate a cartoon sticker pack based on an in-the-wild portrait image. 
To achieve this, we initially perform data augmentation, incorporating processes such as cropping and homography, on the real input image. These augmented data are then employed to train a DreamBooth~\cite{ruiz2023dreambooth}. Subsequently, the trained DreamBooth is utilized to generate a cartoonized portrait image of the subject, guided by a meticulously crafted text prompt. Finally, our model is applied to the cartoonized image to produce outputs featuring four distinct trained expressions.
Please note, directly utilizing DreamBooth to generate images with various expressions doesn't yield satisfactory results due to the  layout change and overfitting issues.
As shown in Figure~\ref{fig:baseline_comparison_costume} (bottom 4 rows), training-free baselines outperform their training-based counterparts. 
This is because the training-based baselines are not robust enough to handle imperfect training pairs.
In contrast, our method outperforms all baselines in both editing fidelity and the preservation of the subject's features, while maintaining high image quality.



\noindent \textbf{User Study}:
We conducted a user study on two real-world applications, each with 12 examples.
Participants were presented with inputs and outputs generated by DiffEdit, SDEdit 0.5, SPADE, BBDM, and our proposed pipeline, randomly shuffled.  
The 32 participants were asked to give a rating from 1 to 5 (higher means better) for each output. 
We normalized the rating of each example and user to remove the user bias. 
In costume editing task, our method achieves the highest average rating, surpassing DiffEdit by 3.3 times, SDEdit 0.5 by 1.8 times, SPADE by 2.1 times, and BBDM by 2.5 times. Similarly, for the expression editing, our method receives the best rating, outperforming DiffEdit by 1.7 times, SDEdit 0.5 by 1.4 times, SPADE by 2.9 times, and BBDM by 1.6 times.
These results demonstrate that our method consistently produces superior visual outcomes compared with baselines in both tasks.

\noindent \textbf{Comparison on Validation Set}:
For quantitative evaluation, we create a validation dataset for each task by generating 1,000 image pairs in two distinct ways. The first approach involves generating paired data following the same methodology described before, resulting in 100 pairs. 
For the second method, we adopt a different strategy aimed at introducing subjects not present in the FFHQ dataset. 
We exclude identity embeddings and add detailed text descriptions of individuals (generated by ChatGPT) to $p$, $p_a$, and $p_b$.
This  yields an additional 900 pairs for evaluation.
We believe a more comprehensive evaluation can be conducted by combining these two types of pairs.
Figure \ref{fig:val_baseline_comparison_expression} and Table \ref{table:baseline_comparison} show that our method outperforms all tested baselines.



\noindent \textbf{Ablation Study}:
We conduct experiments to assess the effectiveness of each component of our model, resulting in four variants: (1) \textit{Ours w/o Prt}, training our model from scratch, (2) \textit{Ours w/o Spt}, removing spatial embeddings $c_s$, (3) \textit{Ours w/o Iemb}, excluding the image embeddings $c_{im}$, and (4) \textit{Ours w/o mask}, eliminating mask guidance during inference. We did not test variants without text conditions since we trained 4 editing directions using one model in the evaluation, and text conditions are used to determine which types of editing to perform at the test time.
As discussed in Section \ref{sec:pipeline}, Table \ref{table:baseline_comparison}, Figure \ref{fig:costume_cond_ablation}, and Figure \ref{fig:mask_ablation} show that our final design outperforms these variants.


\noindent \textbf{Limitations and Future Work}:
The dataset generation strategy assumes Stable Diffusion can generate images in the source and target domains, which might not always be the case. The editing performance is compromised when handling datasets with most of the pairs with significant noise, such as substantial layout and identity differences. 

In the future, we will (1) move away from the constraint of paired data and explore methods for handling unpaired data effectively, (2) reduce the required amount of training data, making the pipeline more efficient and scalable.

\section{Conclusion}

In this paper, we aim for portrait editing such as changing costumes and expressions while preserving the untargeted features. 
We introduce a novel multi-conditioned diffusion model, trained on training pairs generated by our proposed dataset generation strategy. During inference, our model produces an editing mask and uses it to further preserve details of subject features.   Our results on two editing tasks demonstrates superiority over existing state-of-the-art methods both quantitatively and qualitatively.

\noindent\textbf{Societal Impact:} Our method should be used properly and carefully, as it could create fake images, which is an issue with image editing approaches.

\noindent \textbf{Disclaimer}:
If any of the images belongs to you and you would like it removed from the paper, please kindly inform us and provide the relevant evidence, we will update the Arxiv paper to exclude your image.

{
    \small
    \bibliographystyle{ieeenat_fullname}
    \bibliography{main}
}


\end{document}


\maketitlesupplementary


\section{Implementation Details}


\subsection{Paired Data Generation}
\label{sec:data}

Different pretrained Stable Diffusion models are employed for generating two  datasets. The costume editing dataset is created using RealisticVision v4.0~\cite{Civitairv40}. For the cartoon expression editing dataset, fine-tuning is performed on RealisticVision v1.3~\cite{Civitairv13} using 400 images generated by the Samaritan 3D cartoon pretrained Stable Diffusion model~\cite{Civitaisam10}. The fine-tuned model is then used as base model to generate the cartoon expression editing dataset.

For text prompt, we set $p$ to  ``the same [X] on the left and right'', where  [X] corresponds to the gender (either man or woman) detected from the portrait image used to extract identity information. 
$p_a$ is ``a [X], [Y]'', where [Y] refers to the description of the source domain (\ie, normal costume or neutral expression).
$p_b$ is ``a [X], [Z]'', where [Z] is the description of the target domain. In costume editing task, [Z] are ``cute flower costume'', ``cute white sheep costume'', ``Santa Claus costume'', or ``traditional golden palace costume''.  In expression editing task, [Z] are ``angry'', ``shocked'', ``laughing'', or ``crying''. 
In addition, we set $s'_d=0.1$ and $s'_a=s'_b=0.9$.

For the CLIP-based identity encoder, we train it on the FFHQ dataset~\cite{kazemi2014one}. As a variant of the image encoder and global mapper in~\cite{wei2023elite}, our architecture consists of a CLIP image encoder~\cite{radford2021learning} followed by a Decoder layer~\cite{vaswani2017attention} and a multi-layer perceptron.  
This identity encoder translates an input image to  multiple textual word embeddings, which can be regarded as ``new words" in the textual word embedding space.  The resultant embeddings $c_{id}$  have a dimension of $2 \times 768$, which can be regarded as using two words to describe the identity.   
Finally, $c_{id}$ is inserted into the position between ``[X]'' and its previous word in  $c_{p_a}$. In other words, the ``new words'' are inserted right before ``[X]''.   Same operation is also performed for $c_p$ and $c_{p_b}$.
Images from the FFHQ dataset are utilized for identity information extraction due to the dataset's broad diversity of identities across its entirety.

The pose image is generated using Stable Diffusion with the text prompt ``a person, head shot'', resulting in 1000 candidate poses. From these candidates, one pose is randomly selected to serve as the pose condition for generating a pair.

For both datasets, the image size is set to $H=W=512$. The denoising timestep is set to $T=20$, and the guidance scale of the classifier-free guidance is set to $7.5$. We use Stochastic Karras VE scheduler~\cite{Karras2022edm} for denosing.

\subsection{Training Multi-Conditioned Diffusion Model}

In the Multi-Conditioned Diffusion Model (MCDM), we implement $MLP(\cdot)$ following the structure described in the adapter in \cite{karras2023dreampose}. During training, the weights in the first layer of the U-Net related to spatial embeddings $c_s$ are zero-initialized. For other weights, we use the same pretrained Stable Diffusion model, which was adopted for paired data generation, as the base model for initialization.

The training setup includes a batch size of 4 and a learning rate of 5e-6. All models are trained for 200,000 steps, and the training process takes approximately 3 days on a single A100 GPU.

\subsection{Mask-Guided Editing using Trained Model}

During the inference phase, we set the timestep $T$ to 20 and utilize the UniPC scheduler~\cite{zhao2023unipc} for all tasks. Regarding the guidance scale for the classifier-free guidance, we configure $s_1=3$, $s_2=3$, and $s_3=5$. Here, $s_1$, $s_2$, and $s_3$ correspond to image embeddings, input embeddings, and text embeddings, respectively.

For detailed implementation of mask computation,  we first add random noise to the input image $x_A$ for 10 steps using the UniPC scheduler, resulting in $z_{t}$.
Subsequently, we perform one-step denoising processes twice using the trained model, each with a distinct prompt -- $p_a$ and $p_b$. This yields $z'_{t-1}$ (for reconstruction) and $z_{t-1}$ (for target editing), respectively.
The distinction between $z_{t-1}$ and $z'_{t-1}$ is interpretable as the contrast between applying editing and not applying it. This difference is then utilized to compute the editing mask. Noteworthy is the use of $p_a$ for reconstruction, which, as opposed to utilizing a user-defined source prompt in DiffEdit, ensures exact reconstruction and facilitates superior mask computation.
Now, we can compute a binary editing mask $M$ by $N(|z_{t-1} - z'_{t-1}|) > \lambda$, 
where regions with a mask value of 1 indicate areas related for editing, while those with a value of 0 remain unaltered.  $\lambda$ serves as a constant threshold, where a larger value results in fewer regions being marked for editing. We set $\lambda$ to 0.2 and 0.35 for costume and cartoon expression editing, respectively.  The operation $N(\cdot)$ denotes the normalization of the values to a range from 0 to 1, followed by a Gaussian blur with a kernel size of 5. 

In practice, following the approach of DiffEdit~\cite{couairon2022diffedit}, the above process is repeated with a set of 10 different $z_{t-1}$ and $z'_{t-1}$ (generated with different random seeds) to enhance the stability of the mask computation. This helps mitigate the potential impact of variations introduced by different random seeds during the noise addition step.

During inference, we set the timestep $T$ to 20 and employ the UniPC scheduler~\cite{zhao2023unipc} for all tasks. For the guidance scale of the classifier-free guidance, we set $s_1=3$, $s_2=3$, and $s_3=5$.
We set $\lambda$ to 0.2 and 0.35 for costume and cartoon expression editing, respectively.

For each task, we use the same pretrained Stable Diffusion model adopted for dataset generation as the base model for  training-free baselines and initialization of our model.

\section{Experiments}

\subsection{Datasets}

For dataset generation details, please refer to Section \ref{sec:data} for details.

\subsection{Real-World Applications}

\begin{figure*}
    \centering
    \includegraphics[scale=1.3]{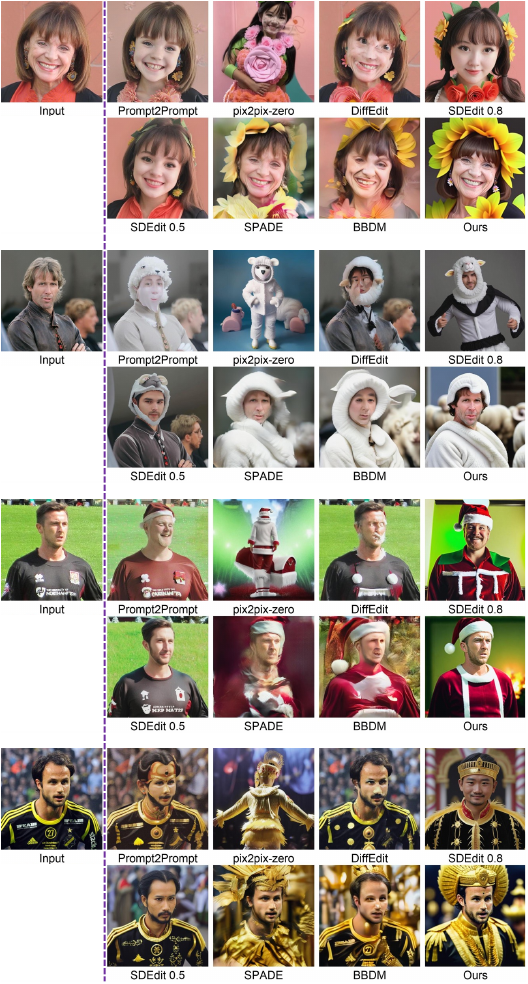}
    \caption{Costume editing comparison with baselines on in-the-wild images. Editing directions from top to bottom are to flower, sheep, Santa, and royal costume. 
    Our method produces high-quality editing results while preserving the subject features. }
    \label{fig:test_costume1}
\end{figure*}

\begin{figure*}
    \centering
    \includegraphics[scale=1.3]{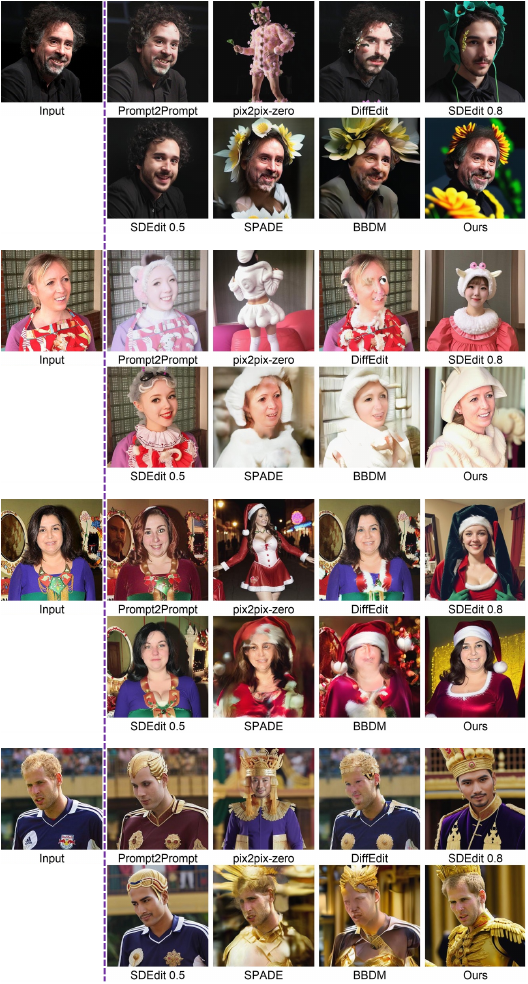}
    \caption{Costume editing comparison with baselines on in-the-wild images. Editing directions from top to bottom are to flower, sheep, Santa, and royal costume. 
    Our method produces high-quality editing results while preserving the subject features. }
    \label{fig:test_costume2}
\end{figure*}

\begin{figure*}
    \centering
    \includegraphics[scale=1.3]{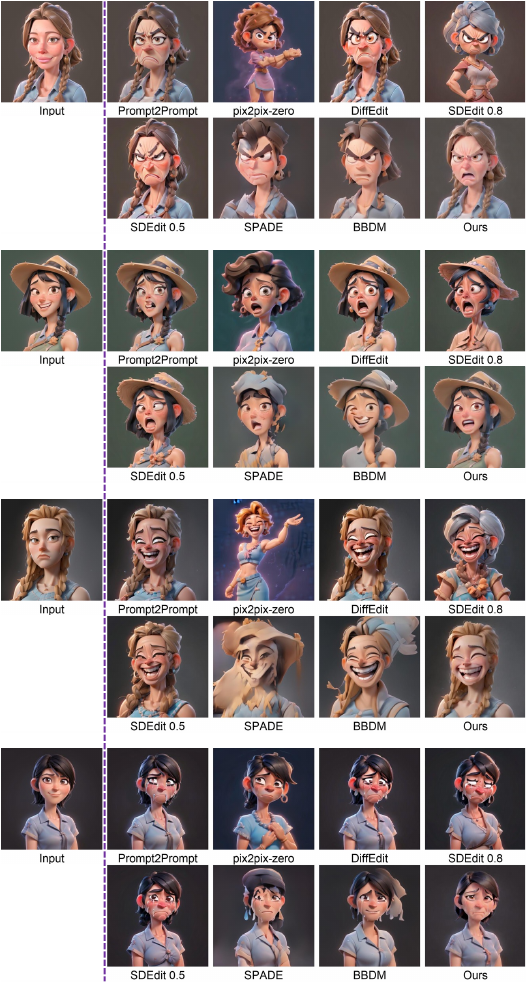}
    \caption{Expression editing comparison with baselines on cartoon input produced by DreamBooth. Editing directions from top to bottom are to angry, shocked, laughing, and crying expression. 
    Our method produces high-quality editing results while preserving the subject features. }
    \label{fig:test_exp1}
\end{figure*}

\begin{figure*}
    \centering
    \includegraphics[scale=1.3]{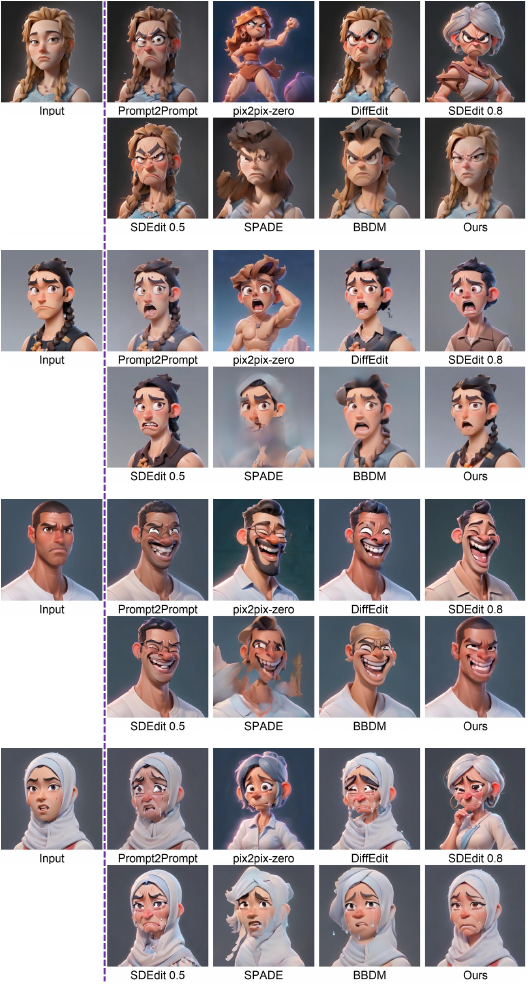}
    \caption{Expression editing comparison with baselines on cartoon input produced by DreamBooth. Editing directions from top to bottom are to angry, shocked, laughing, and crying expression. 
    Our method produces high-quality editing results while preserving the subject features. }
    \label{fig:test_exp2}
\end{figure*}

Figure \ref{fig:test_costume1} and Figure \ref{fig:test_costume2} show more results of costume editing task.   
Figure \ref{fig:test_exp1} and Figure \ref{fig:test_exp2} showcase more expression editing results. 
In summary, our method outperforms all baselines in both editing fidelity and the preservation of the subject's features, while maintaining high image quality.

\subsection{Comparison on Validation Set}

\begin{figure*}
    \centering
    \includegraphics[scale=1.3]{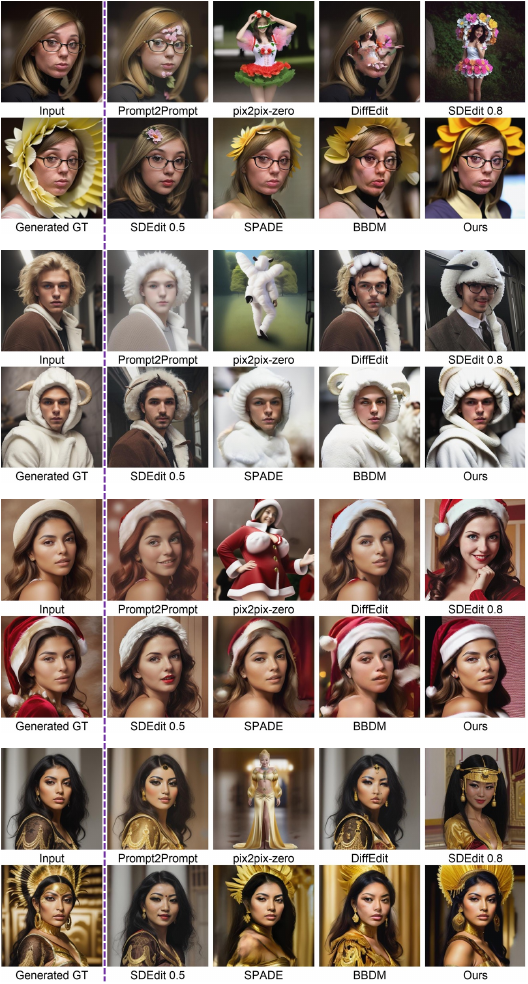}
    \caption{Costume editing comparison on validation set. Editing directions from top to bottom are to flower, sheep, Santa, and royal costume. 
    Our method produces high-quality editing results while preserving the subject feature. Note that validation set input is from generated pairs, so baseline results look better than those in real-world applications.}
    \label{fig:val_costume}
\end{figure*}

\begin{figure*}
    \centering
    \includegraphics[scale=1.3]{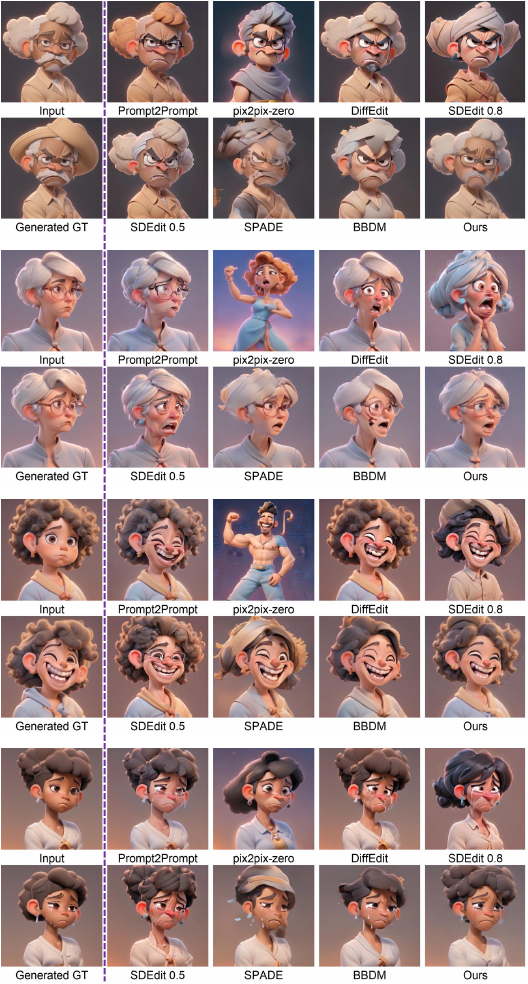 }
    \caption{Expression editing comparison on validation set. Editing directions from top to bottom are to angry, shocked, laughing, and crying expression. 
    Our method produces high-quality editing results while preserving the subject feature. Note that validation set input is from generated pairs, so baseline results look better than those in real-world applications.}
    \label{fig:val_exp}
\end{figure*}

Figure \ref{fig:val_costume} and Figure \ref{fig:val_exp} show the comparison between our method and all baselines on costume editing and expression editing dataset, respectively. Our method outperforms all baselines in generating high-quality editing outputs. 

\subsection{Ablation Study}

\begin{figure*}[!t]
\begin{center}
\begin{minipage}{0.12\linewidth}
\vspace{-60pt}
\begin{tabular}{c}
\quad Flower
\end{tabular}
\end{minipage}
\begin{subfigure}[b]{0.14\linewidth}
\includegraphics[width=\linewidth]{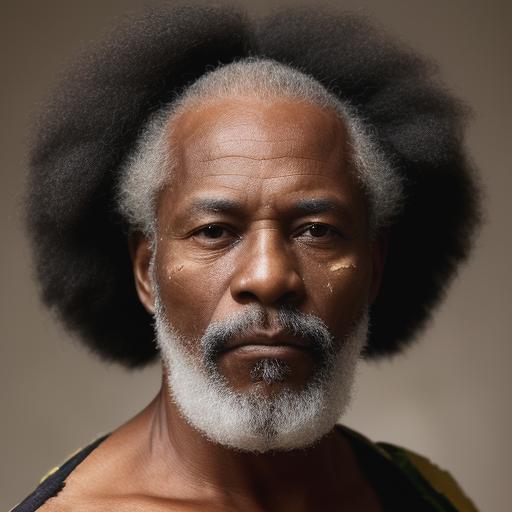} %
\end{subfigure}
\begin{subfigure}[b]{0.14\linewidth}
\includegraphics[width=\linewidth]{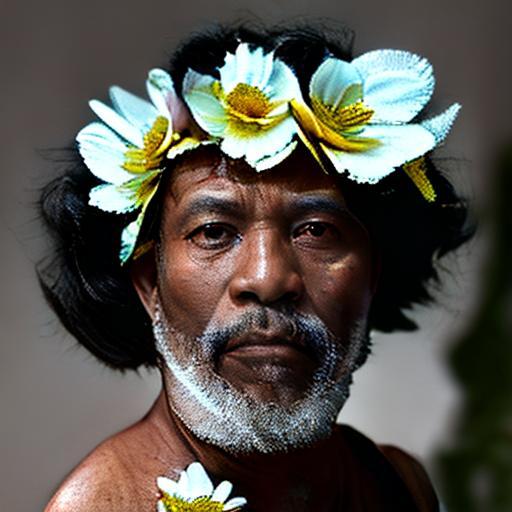} %
\end{subfigure}
\begin{subfigure}[b]{0.14\linewidth}
\includegraphics[width=\linewidth]{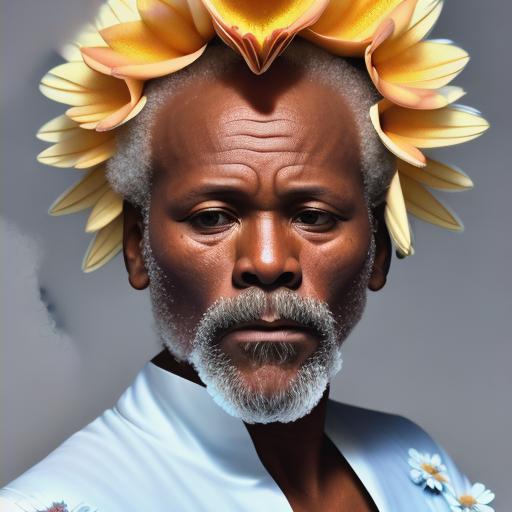} %
\end{subfigure}
\begin{subfigure}[b]{0.14\linewidth}
\includegraphics[width=\linewidth]{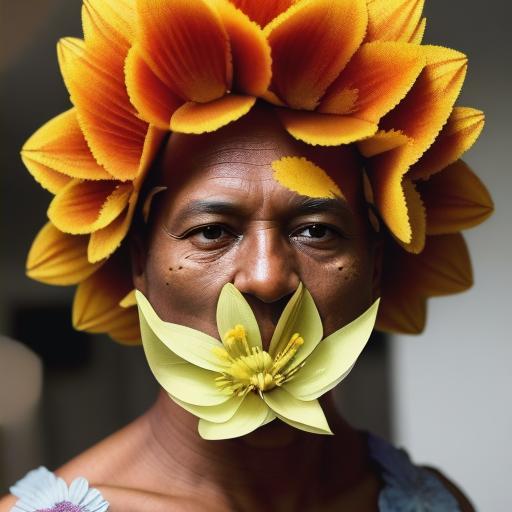} %
\end{subfigure}
\begin{subfigure}[b]{0.14\linewidth}
\includegraphics[width=\linewidth]{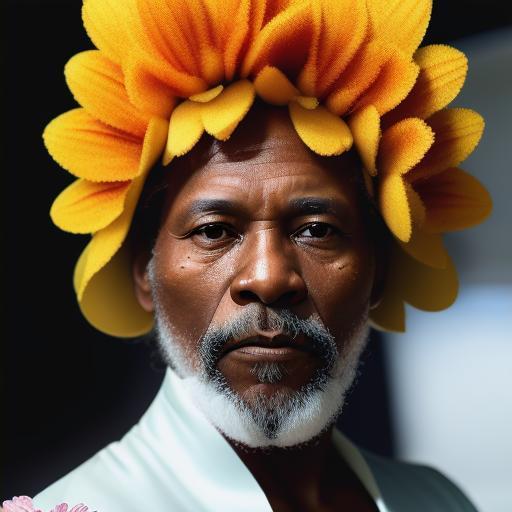} %
\end{subfigure}
\begin{subfigure}[b]{0.14\linewidth}
\includegraphics[width=\linewidth]{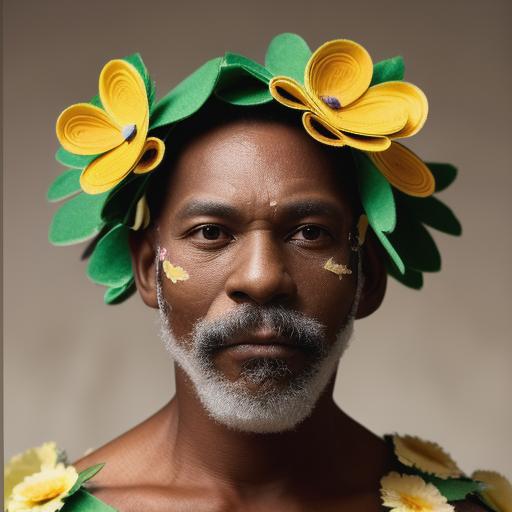} %
\end{subfigure}
\\
\begin{minipage}{0.12\linewidth}
\vspace{-60pt}
\begin{tabular}{c}
\quad Sheep 
\end{tabular}
\end{minipage}
\begin{subfigure}[b]{0.14\linewidth}
\includegraphics[width=\linewidth]{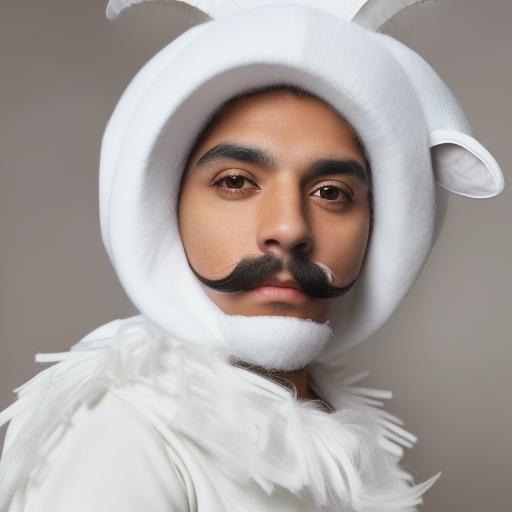} %
\end{subfigure}
\begin{subfigure}[b]{0.14\linewidth}
\includegraphics[width=\linewidth]{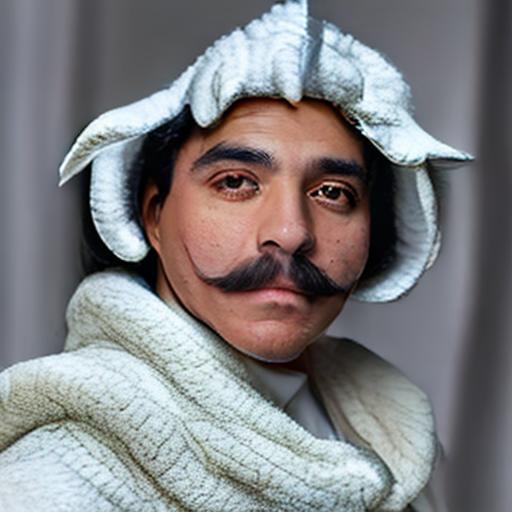} %
\end{subfigure}
\begin{subfigure}[b]{0.14\linewidth}
\includegraphics[width=\linewidth]{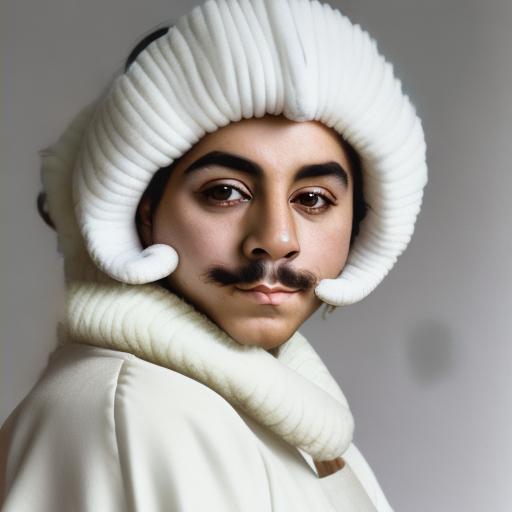} %
\end{subfigure}
\begin{subfigure}[b]{0.14\linewidth}
\includegraphics[width=\linewidth]{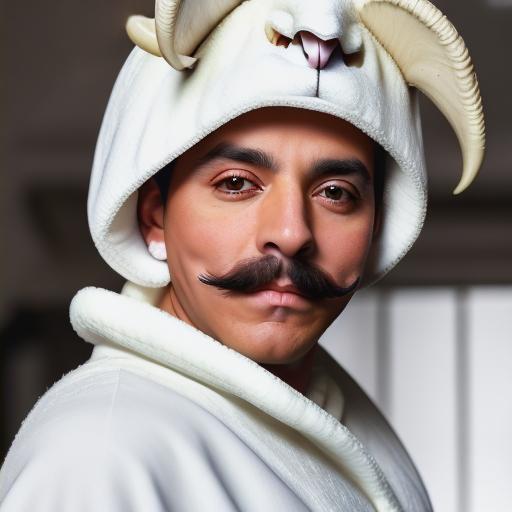} %
\end{subfigure}
\begin{subfigure}[b]{0.14\linewidth}
\includegraphics[width=\linewidth]{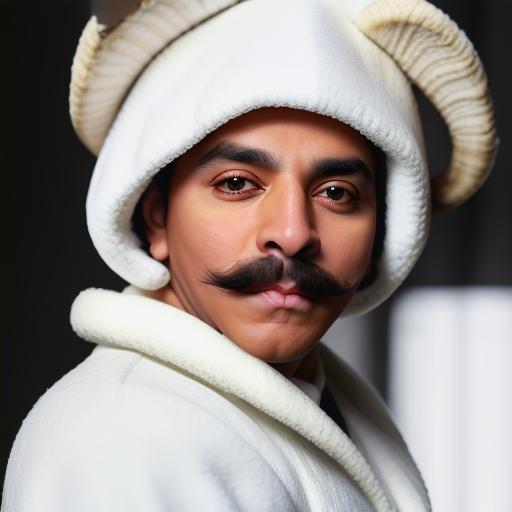} %
\end{subfigure}
\begin{subfigure}[b]{0.14\linewidth}
\includegraphics[width=\linewidth]{supp_fig/ablation/costume/32_1/GT.jpg} %
\end{subfigure}
\\
\begin{minipage}{0.12\linewidth}
\vspace{-60pt}
\begin{tabular}{c}
\quad Santa \\ \quad Claus
\end{tabular}
\end{minipage}
\begin{subfigure}[b]{0.14\linewidth}
\includegraphics[width=\linewidth]{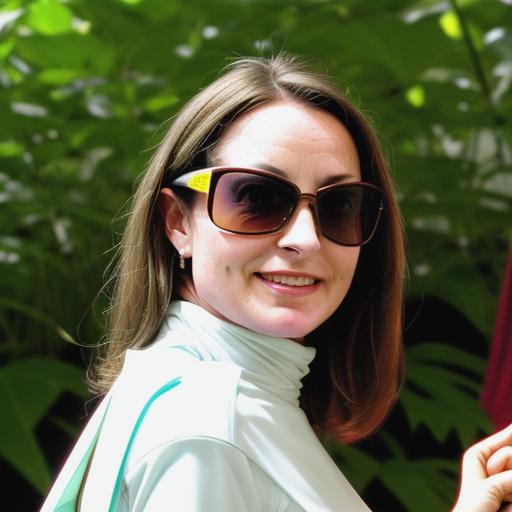} %
\end{subfigure}
\begin{subfigure}[b]{0.14\linewidth}
\includegraphics[width=\linewidth]{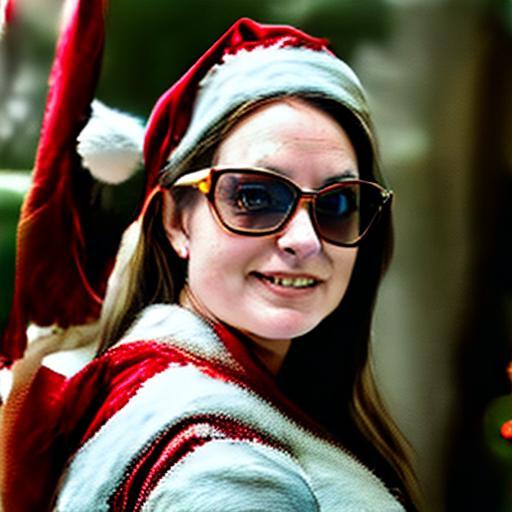} %
\end{subfigure}
\begin{subfigure}[b]{0.14\linewidth}
\includegraphics[width=\linewidth]{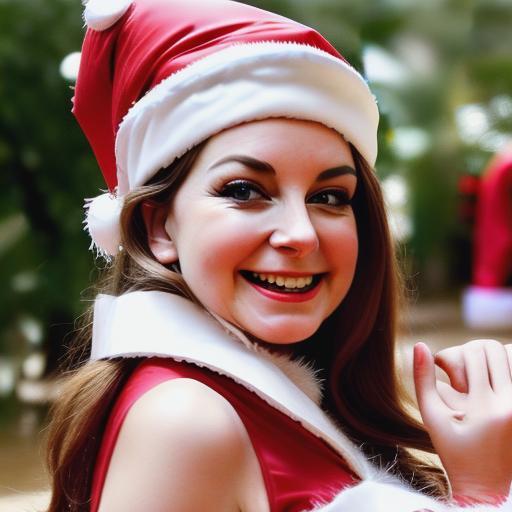} %
\end{subfigure}
\begin{subfigure}[b]{0.14\linewidth}
\includegraphics[width=\linewidth]{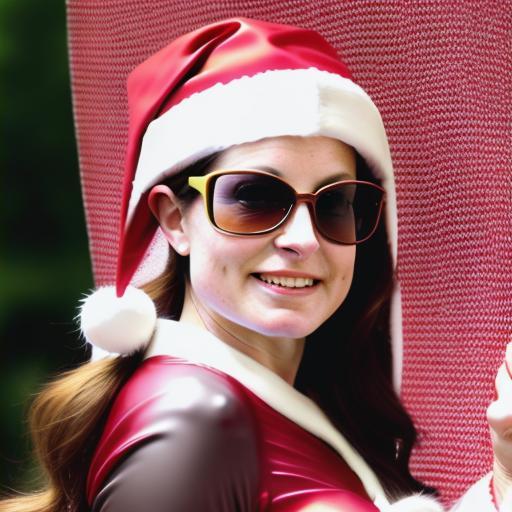} %
\end{subfigure}
\begin{subfigure}[b]{0.14\linewidth}
\includegraphics[width=\linewidth]{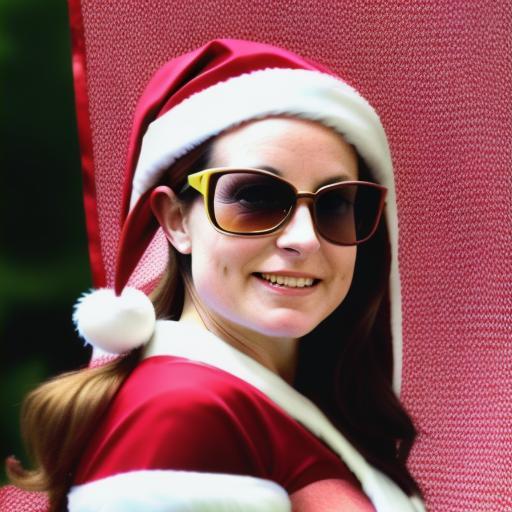} %
\end{subfigure}
\begin{subfigure}[b]{0.14\linewidth}
\includegraphics[width=\linewidth]{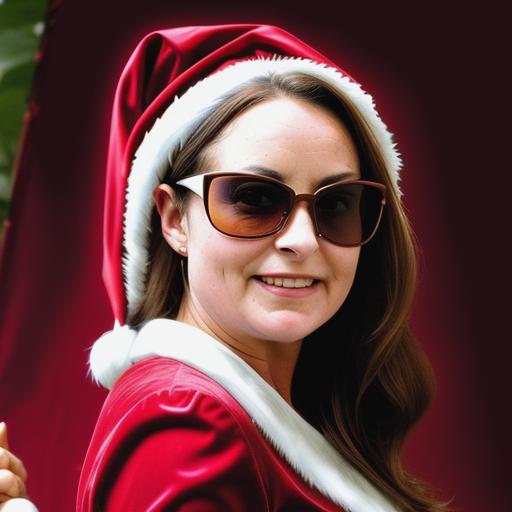} %
\end{subfigure}
\\
\begin{minipage}{0.12\linewidth}
\vspace{-60pt}
\begin{tabular}{c}
\quad Royal 
\end{tabular}
\end{minipage}
\begin{subfigure}[b]{0.14\linewidth}
\includegraphics[width=\linewidth]{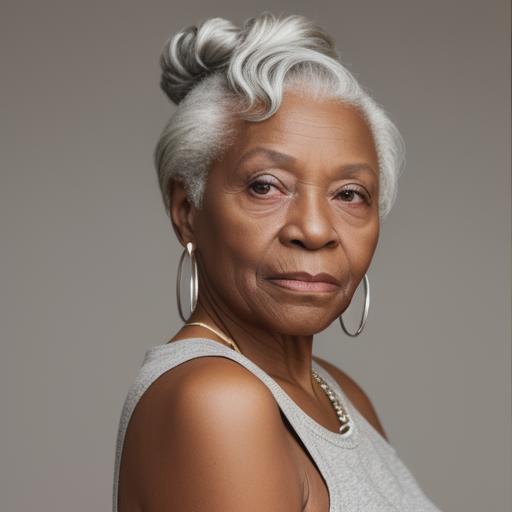} %
\end{subfigure}
\begin{subfigure}[b]{0.14\linewidth}
\includegraphics[width=\linewidth]{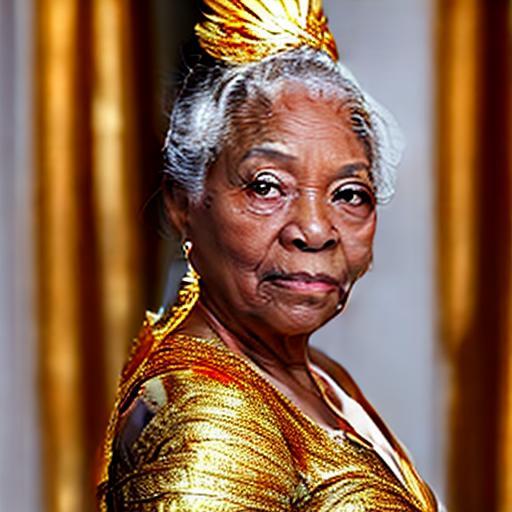} %
\end{subfigure}
\begin{subfigure}[b]{0.14\linewidth}
\includegraphics[width=\linewidth]{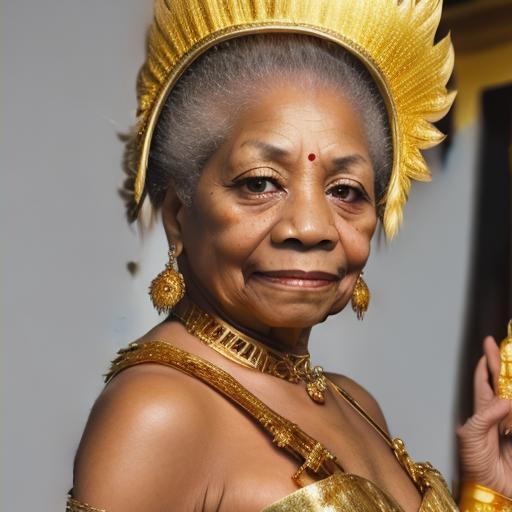} %
\end{subfigure}
\begin{subfigure}[b]{0.14\linewidth}
\includegraphics[width=\linewidth]{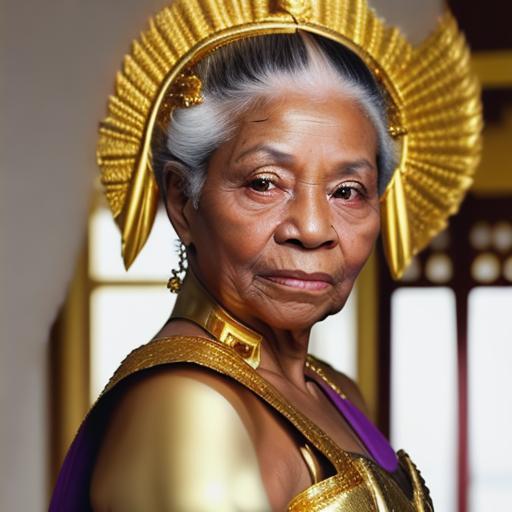} %
\end{subfigure}
\begin{subfigure}[b]{0.14\linewidth}
\includegraphics[width=\linewidth]{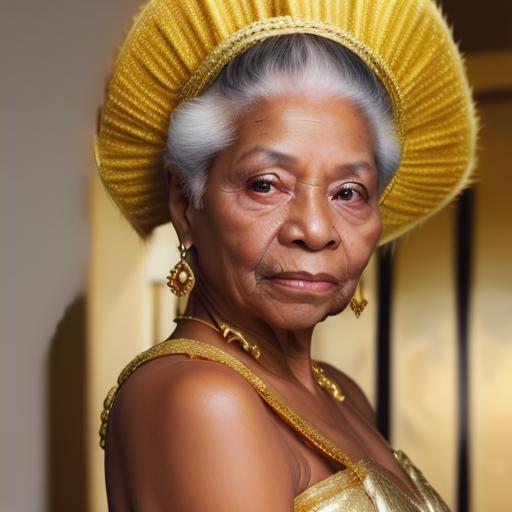} %
\end{subfigure}
\begin{subfigure}[b]{0.14\linewidth}
\includegraphics[width=\linewidth]{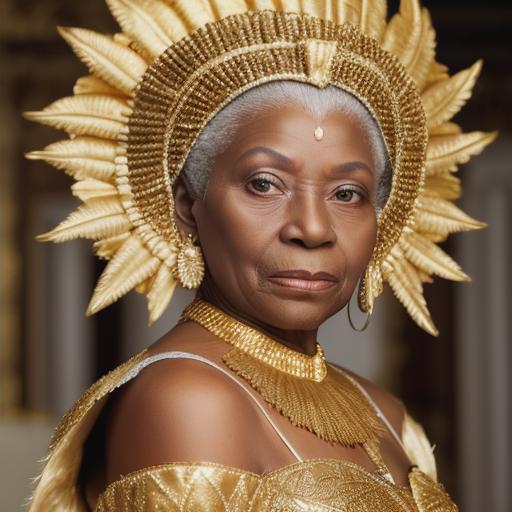} %
\end{subfigure}
\\
\begin{minipage}{0.12\linewidth}
\vspace{-60pt}
\begin{tabular}{c}
\quad Angry
\end{tabular}
\end{minipage}
\begin{subfigure}[b]{0.14\linewidth}
\includegraphics[width=\linewidth]{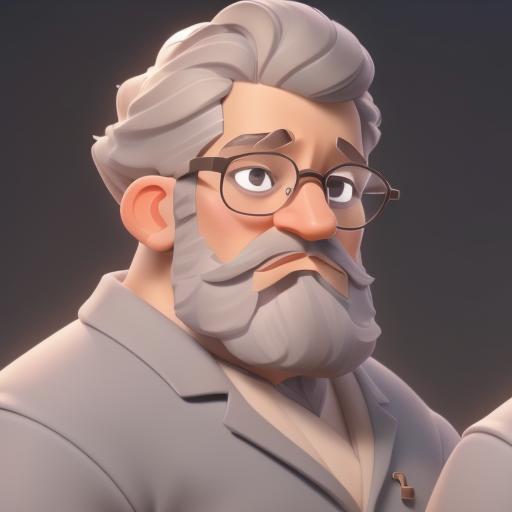} %
\end{subfigure}
\begin{subfigure}[b]{0.14\linewidth}
\includegraphics[width=\linewidth]{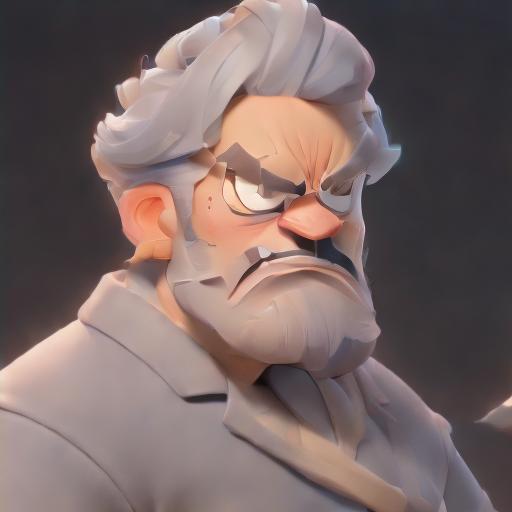} %
\end{subfigure}
\begin{subfigure}[b]{0.14\linewidth}
\includegraphics[width=\linewidth]{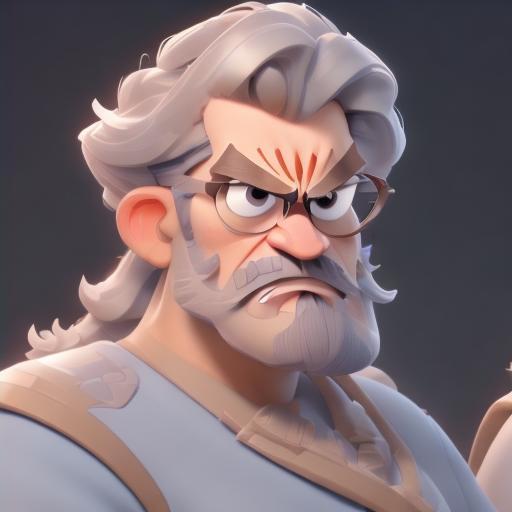} %
\end{subfigure}
\begin{subfigure}[b]{0.14\linewidth}
\includegraphics[width=\linewidth]{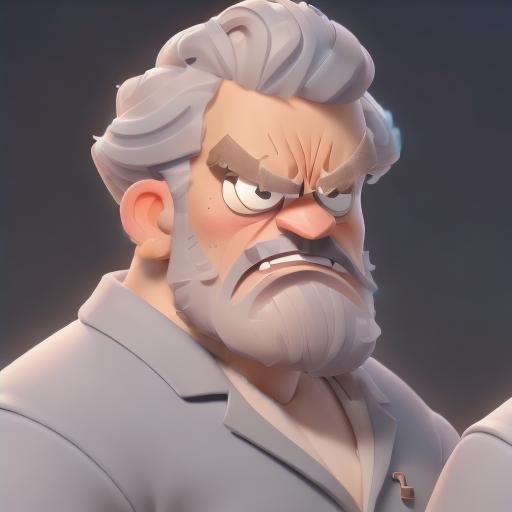} %
\end{subfigure}
\begin{subfigure}[b]{0.14\linewidth}
\includegraphics[width=\linewidth]{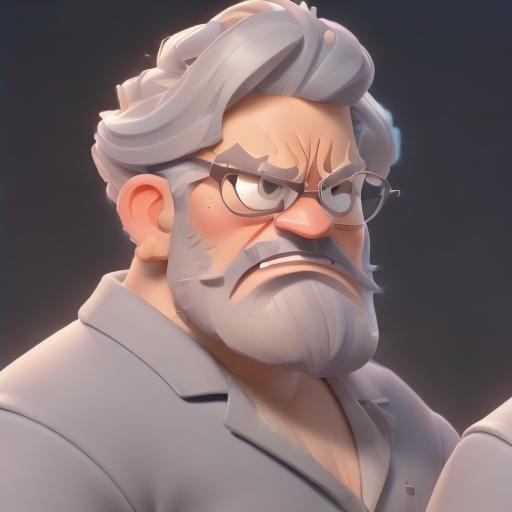} %
\end{subfigure}
\begin{subfigure}[b]{0.14\linewidth}
\includegraphics[width=\linewidth]{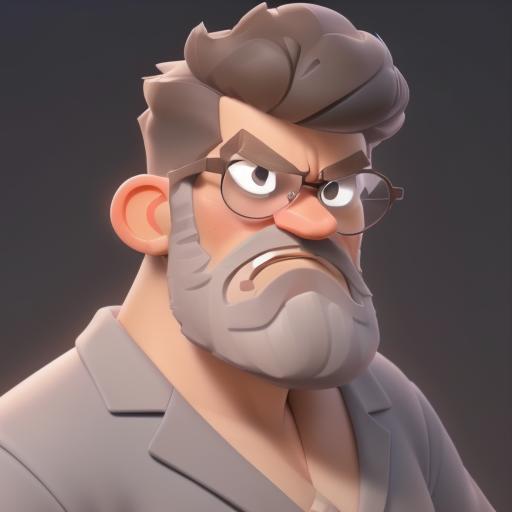} %
\end{subfigure}
\\
\begin{minipage}{0.12\linewidth}
\vspace{-60pt}
\begin{tabular}{c}
\quad Shocked
\end{tabular}
\end{minipage}
\begin{subfigure}[b]{0.14\linewidth}
\includegraphics[width=\linewidth]{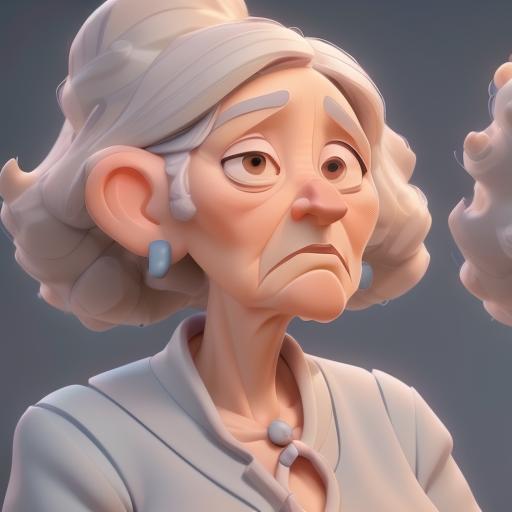} %
\end{subfigure}
\begin{subfigure}[b]{0.14\linewidth}
\includegraphics[width=\linewidth]{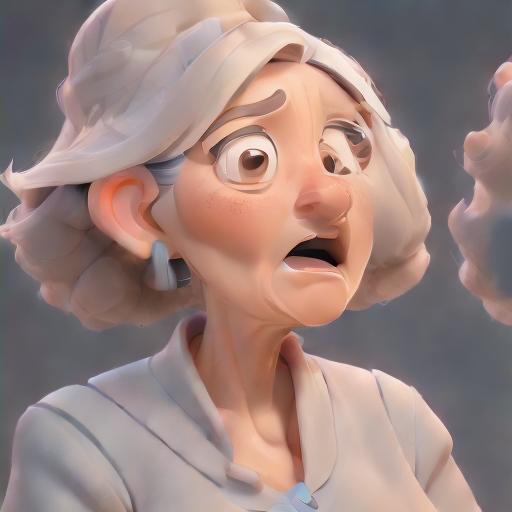} %
\end{subfigure}
\begin{subfigure}[b]{0.14\linewidth}
\includegraphics[width=\linewidth]{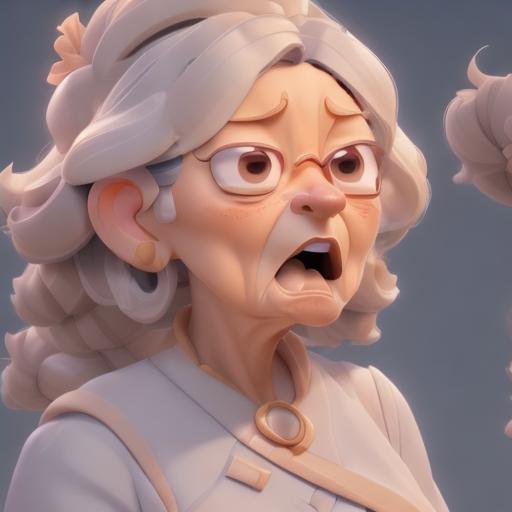} %
\end{subfigure}
\begin{subfigure}[b]{0.14\linewidth}
\includegraphics[width=\linewidth]{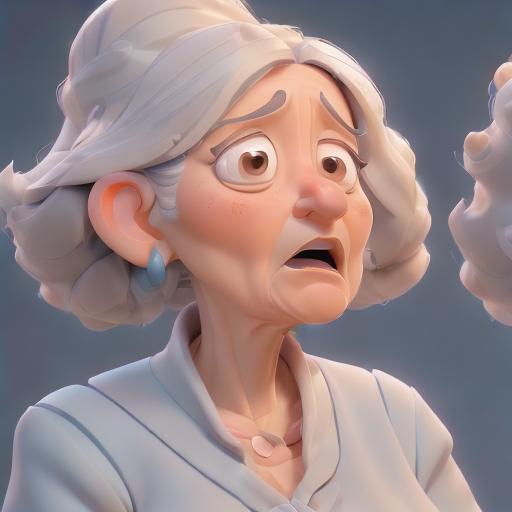} %
\end{subfigure}
\begin{subfigure}[b]{0.14\linewidth}
\includegraphics[width=\linewidth]{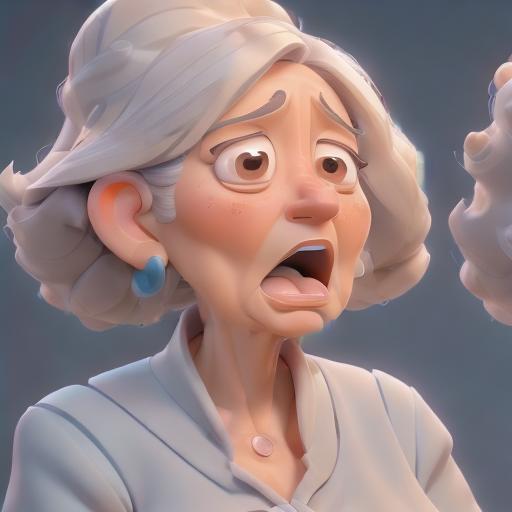} %
\end{subfigure}
\begin{subfigure}[b]{0.14\linewidth}
\includegraphics[width=\linewidth]{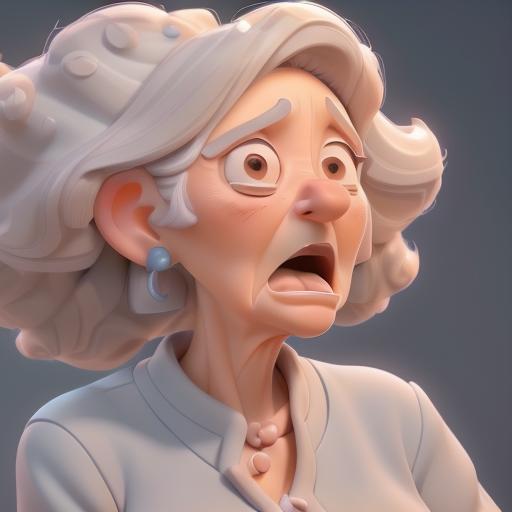} %
\end{subfigure}
\\
\begin{minipage}{0.12\linewidth}
\vspace{-60pt}
\begin{tabular}{c}
\quad Laughing
\end{tabular}
\end{minipage}
\begin{subfigure}[b]{0.14\linewidth}
\includegraphics[width=\linewidth]{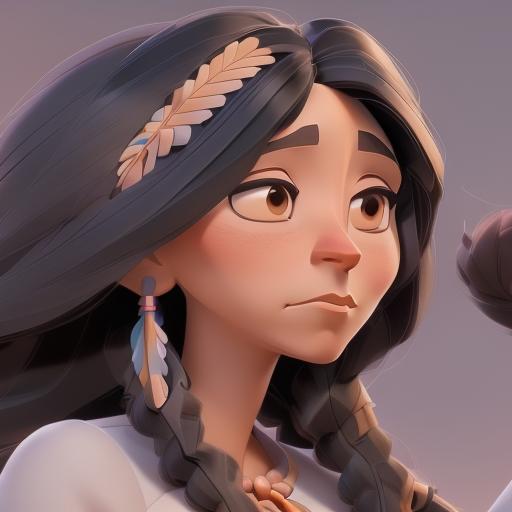} %
\end{subfigure}
\begin{subfigure}[b]{0.14\linewidth}
\includegraphics[width=\linewidth]{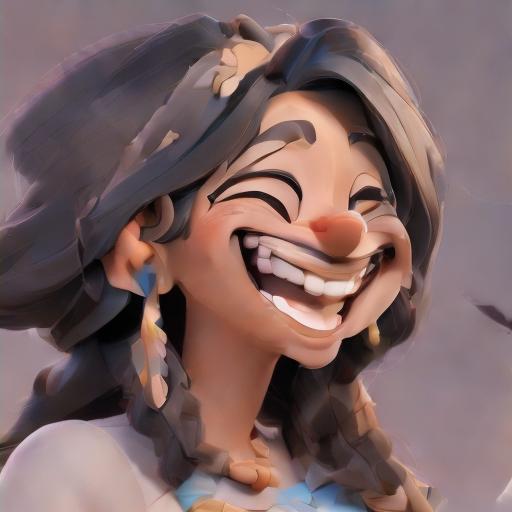} %
\end{subfigure}
\begin{subfigure}[b]{0.14\linewidth}
\includegraphics[width=\linewidth]{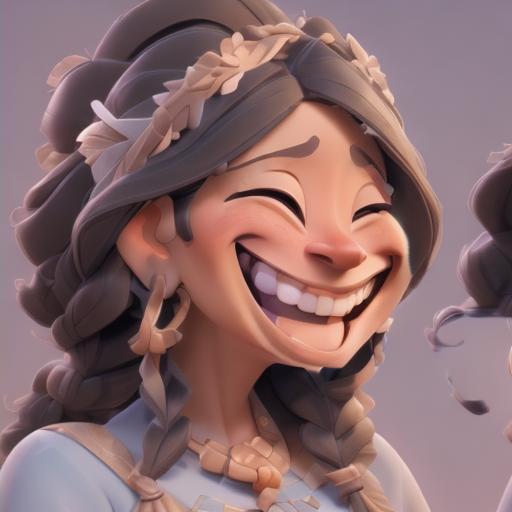} %
\end{subfigure}
\begin{subfigure}[b]{0.14\linewidth}
\includegraphics[width=\linewidth]{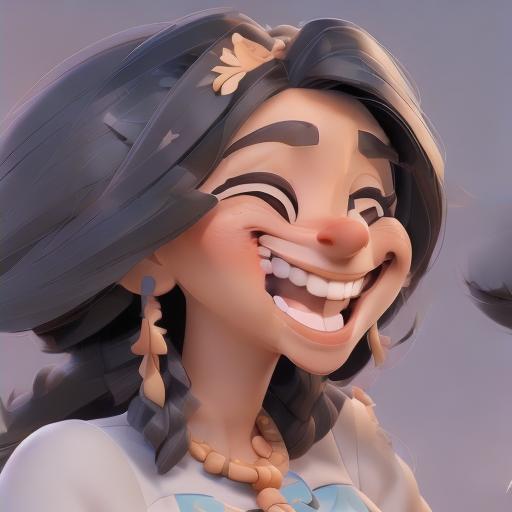} %
\end{subfigure}
\begin{subfigure}[b]{0.14\linewidth}
\includegraphics[width=\linewidth]{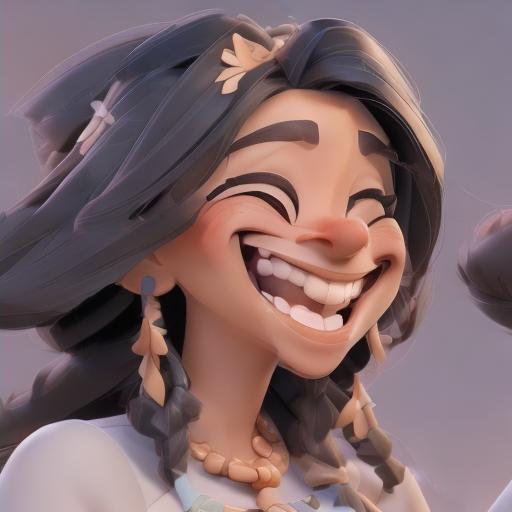} %
\end{subfigure}
\begin{subfigure}[b]{0.14\linewidth}
\includegraphics[width=\linewidth]{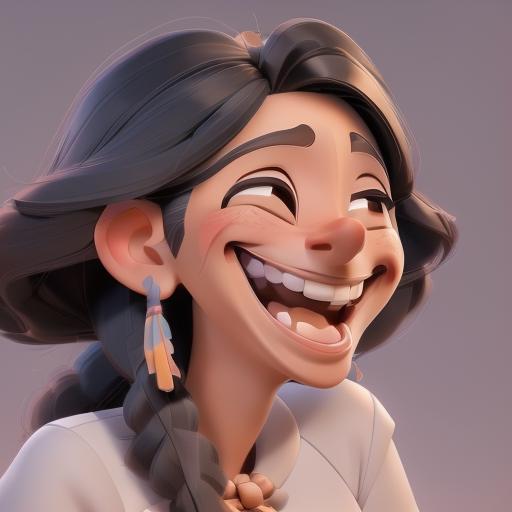} %
\end{subfigure}
\\
\begin{minipage}{0.12\linewidth}
\vspace{-87pt}
\begin{tabular}{c}
\quad Crying
\end{tabular}
\end{minipage}
\begin{subfigure}[b]{0.14\linewidth}
\includegraphics[width=\linewidth]{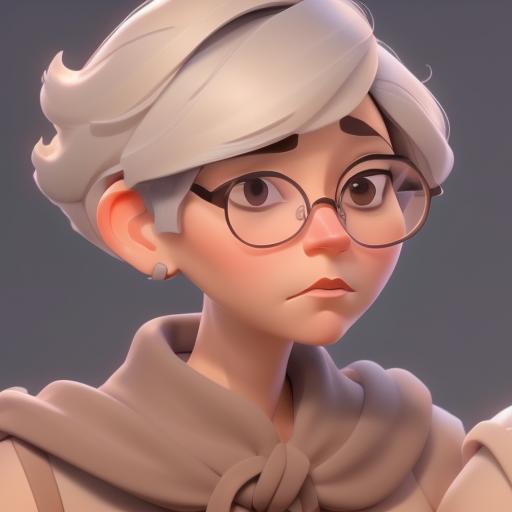} %
\caption{Input}
\end{subfigure}
\begin{subfigure}[b]{0.14\linewidth}
\includegraphics[width=\linewidth]{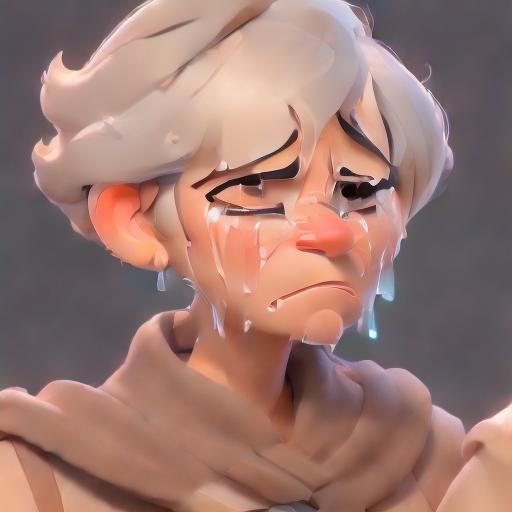} %
\caption{Ours w/o Prt}
\end{subfigure}
\begin{subfigure}[b]{0.14\linewidth}
\includegraphics[width=\linewidth]{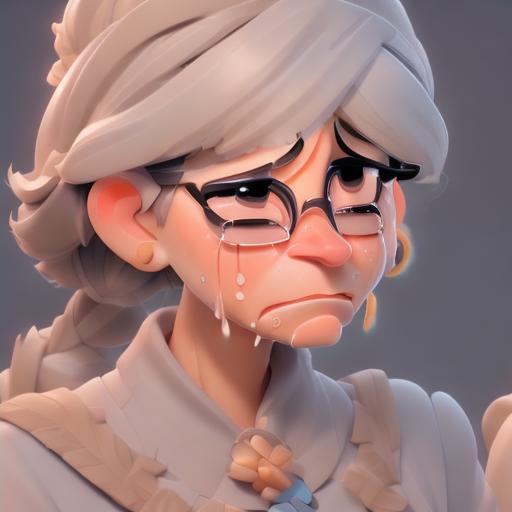} %
\caption{Ours w/o Spt}
\end{subfigure}
\begin{subfigure}[b]{0.14\linewidth}
\includegraphics[width=\linewidth]{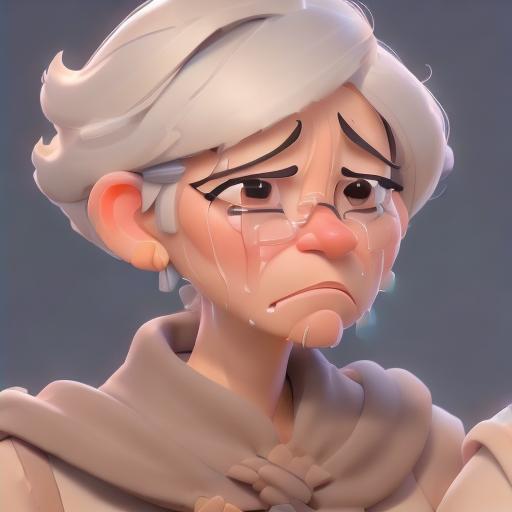} %
\caption{Ours w/o Iemb}
\end{subfigure}
\begin{subfigure}[b]{0.14\linewidth}
\includegraphics[width=\linewidth]{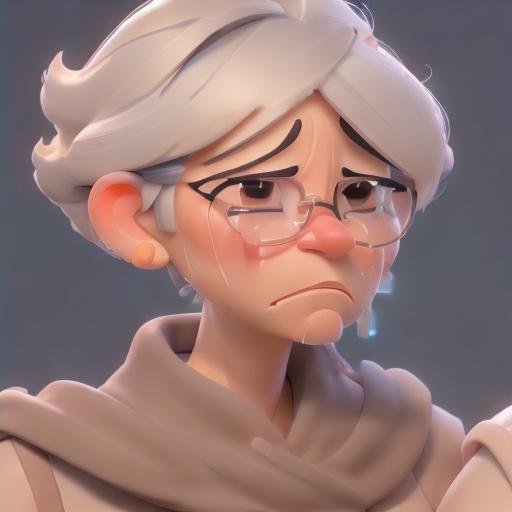} %
\caption{Ours}
\end{subfigure}
\begin{subfigure}[b]{0.14\linewidth}
\includegraphics[width=\linewidth]{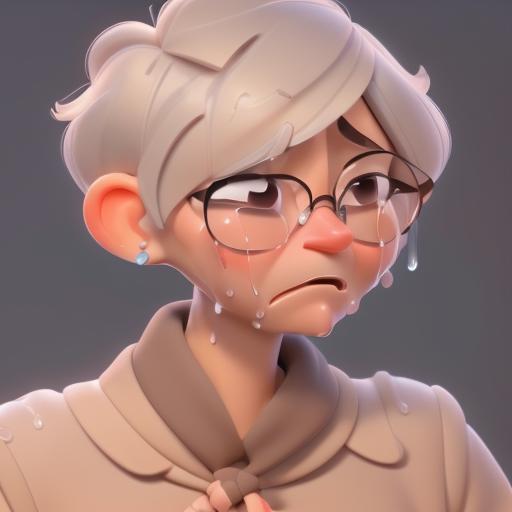} %
\caption{Generated GT}
\end{subfigure}
\caption{
Ablation Study of design choice of different conditions of MCDM. Training from scratch (b) yields the poorest image quality due to the absence of image generation priors and text prompt interpretation. Dropping spatial embeddings (c) fails to preserve spatial layout. Excluding image embeddings (d) fails to learn precise editing direction, leading to artifacts in the outputs. In contrast, our full pipeline (e) produces the best editing results.
}
\label{fig:supp_ablation}
\end{center}
\end{figure*}

Figure \ref{fig:supp_ablation} shows additional results of ablation of different condition for MCDM, demonstrating the effectiveness of our final design.

 \clearpage
{
    \small
    \bibliographystyle{ieeenat_fullname}
    \bibliography{suppl}
}
